\documentclass{article}
\usepackage{iclr2027_conference,times}
\usepackage[utf8]{inputenc}
\usepackage[T1]{fontenc}
\usepackage{hyperref,url,booktabs,amsfonts,amsmath,amssymb,nicefrac,microtype,xcolor,xspace}
\usepackage{float,graphicx,subcaption,multirow,array,tabularx,etoolbox}
\floatstyle{ruled}
\newfloat{algorithm}{tbp}{loa}
\floatname{algorithm}{Algorithm}
\DeclareUnicodeCharacter{00D7}{\ensuremath{\times}}
\DeclareUnicodeCharacter{2192}{\ensuremath{\rightarrow}}
\newcommand{\da}{\textsc{DreamAvoid}\xspace}
\hypersetup{
 pdftitle={DreamAvoid: Critical-Phase Test-Time Dreaming to Avoid Failures in VLA Policies},pdfauthor={Xianzhe Fan, Yuxiang Lu, Shenyuan Gao, Xiaoyang Wu, Ruihua Han, Manling Li, Hengshuang Zhao}}
\title{\da: Critical-Phase Test-Time\\Dreaming to Avoid Failures in VLA Policies}
\author{%
  \textbf{Xianzhe Fan}\textsuperscript{1} \quad
  \textbf{Yuxiang Lu}\textsuperscript{1} \quad
  \textbf{Shenyuan Gao}\textsuperscript{2} \quad
  \textbf{Xiaoyang Wu}\textsuperscript{1} \\
  \textbf{Ruihua Han}\textsuperscript{1} \quad
  \textbf{Manling Li}\textsuperscript{3} \quad
  \textbf{Hengshuang Zhao}\textsuperscript{1} \\
  {\normalfont\small
  \textsuperscript{1}HKU \quad
  \textsuperscript{2}HKUST \quad
  \textsuperscript{3}Northwestern University
  }
}

\iclrfinalcopy

\begin{document}
\maketitle
\lhead{Preprint}
\noindent\begin{minipage}{\textwidth}
    \centering
    \begingroup
      \setkeys{Gin}{width=\textwidth}%
      \includegraphics{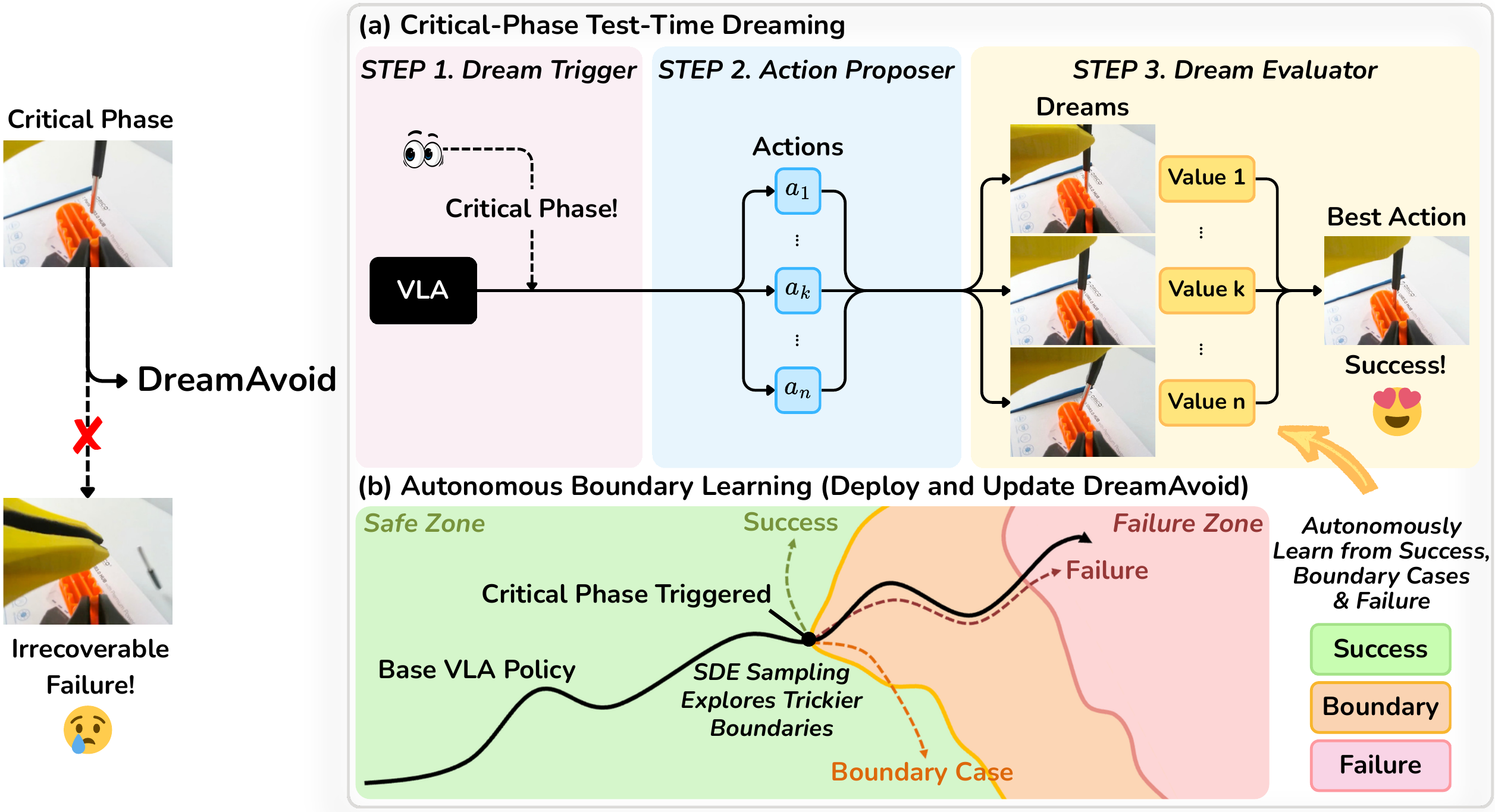}%
      \captionsetup{type=figure}
      \caption{We propose \da, (a) a \textit{critical-phase test-time dreaming} framework that enables the VLA to anticipate and avoid failure. This framework consists of a Dream Trigger, an Action Proposer, and a Dream Evaluator. (b) We also introduce an \textit{autonomous boundary learning} paradigm to equip the world model-based Dream Evaluator with failure awareness.}
      \label{fig:teaser}
    \endgroup
\end{minipage}

\begin{abstract}


Vision-Language-Action (VLA) models are often brittle in fine-grained manipulation, where minor action errors during the critical phases can rapidly escalate into irrecoverable failures. Since existing VLA models rely predominantly on successful demonstrations for training, they lack an explicit awareness of failure during these critical phases. To address this, we propose \da, a critical-phase test-time dreaming framework that enables VLA models to anticipate and avoid failures. We also introduce an autonomous boundary learning paradigm to refine the system's understanding of the subtle boundary between success and failure. Specifically, we (1) utilize a Dream Trigger to determine whether the execution has entered a critical phase, (2) sample multiple candidate action chunks from the VLA via an Action Proposer, and (3) employ a Dream Evaluator, jointly trained on mixed data (success, failure, and boundary cases), to ``dream'' the short-horizon futures corresponding to the candidate actions, evaluate their values, and select the optimal action. We conduct extensive evaluations on real-world manipulation tasks and simulation benchmarks. The results demonstrate that \da can effectively avoid failures, thereby improving the overall task success rate. Across four real-world tasks, \da achieves 72.5\% success, compared with 48.8\% for the base policy and 54.4\% for GPC-RANK. Our code is available at \href{https://github.com/XianzheFan/DreamAvoid}{https://github.com/XianzheFan/DreamAvoid}.

\end{abstract}

\section{Introduction}

Vision-Language-Action (VLA) models provide a promising framework for general-purpose robotic manipulation by directly mapping multimodal observations and language instructions into the action space \citep{kim2025openvla,pmlr-v305-black25a,BlackK-RSS-25}. However, they remain fragile in fine-grained manipulation scenarios. In tasks such as insertion, precise alignment, and contact-sensitive placement, a small local action error can rapidly escalate into an irrecoverable failure \citep{xu2026rl,zhang2025safe,hu2025vlsa}. In many cases, the policy does not perform poorly over the entire trajectory; rather, failures are often triggered by brief yet low-tolerance \textit{critical phases} \citep{onishi2026exploration}.
How can we equip VLAs with stronger consequence-awareness during these critical phases? In human cognition, complex physical tasks are often balanced between fast, intuitive execution and slow, deliberate foresight. When threading a needle or plugging in a cord, we do not consciously plan every minute movement; we rely on intuitive motor priors. However, at critical moments, we briefly pause and simulate in our minds, ``dreaming'' of the physical consequences our actions are about to bring, to avoid making mistakes.

Inspired by this, introducing additional computation during test time to endow VLAs with a ``dreaming'' capability emerges as a natural choice. Powerful large-scale pre-trained world models \citep{gao2026dreamdojo} can help capture complex environmental dynamics and generate high-fidelity future simulations based on candidate actions. Can we leverage this capability to achieve better consequence-awareness for robots through dreaming at test-time? We conducted a pilot study on a charger plugging task, comparing two test-time compute paradigms: (1) Direct Semantic Evaluation, which uses a VLM to predict success or failure by receiving past and current visual observations and future joint-space or end-effector-space numerical action sequences; and (2) Explicit Spatiotemporal Dreaming, which uses a large-scale pre-trained action-conditioned world model to explicitly forward-render actions into future visual frames, which are then evaluated by the same VLM. The results showed that when faced with abstract numerical inputs, the VLM tends towards a conservative ``almost certain to fail'' prediction pattern, whereas its accuracy improves to 80.0\% when evaluating the visual simulations provided by the world model. This contrast reveals a key insight: for fine-grained manipulation, visual consequence prediction can help bridge the gap between abstract actions and physical consequences. Therefore, rendering actions into high-fidelity future frames for explicit test-time dreaming is a promising approach.

Following the idea of introducing extra computation at test-time, several studies have proposed their own solutions. RoboMonkey utilizes a VLM as an online verifier to score and filter candidate actions \citep{kwok2025robomonkey}. GPC and Visual MPC evaluate candidate actions through predicted consequences, while VLA-Reasoner performs online tree search \citep{qi2026inference,finn2017deep,guo2025vla}. On the other hand, some research achieves dynamic scaling of test-time compute through implicit iterative reasoning within a continuous latent space \citep{tur2026recurrentdepth}.
However, implementing test-time computation in robotic manipulation tasks still faces two challenges. (1) Always-on inference can be costly \citep{yang2026seeing}; executing heavy rollouts at every replan incurs additional computation even during routine execution \citep{tur2026recurrentdepth,guo2025vla}. This motivates selective intervention that concentrates computation on consequential decisions while retaining the base policy during routine execution. (2) Existing VLAs typically rely on successful trajectories for training \citep{kim2025openvla}. Such training teaches policies to imitate successful behavior but provides limited supervision on unsuccessful and near-boundary outcomes \citep{BlackK-RSS-25,wu2025learning}. This deficiency is particularly prominent during critical phases, where the robot needs to distinguish between subtle action differences that carry significant consequences.

To address these challenges, we propose \da, a critical-phase test-time dreaming framework designed to enhance the ability of VLAs to avoid failures (Figure \ref{fig:teaser}). We also introduce an autonomous boundary learning paradigm. Specifically,
(1) \da proposes a selective intervention mechanism. During routine execution, the robot follows the frozen base policy. When a lightweight Dream Trigger detects an impending critical phase, the system invokes an Action Proposer to sample multiple diverse candidate action chunks. Subsequently, a world-model-based Dream Evaluator engages in short-horizon dreaming to predict their future visual evolution and values, selecting the highest-scoring candidate.
(2) \da proposes an autonomous boundary learning paradigm, introducing failure and boundary data from which the world models can benefit. For the dreams to truly serve a preventive purpose, the world model must possess the ability to have ``nightmares.'' If trained solely on success data, the world model may generate optimistic hallucinations due to a lack of negative boundary constraints. By deploying \da to explore and collect success, failure, and boundary data, we enable the world model to autonomously learn the boundary between success and failure.

We evaluate \da across multiple real-world tasks that require precise alignment, stable contact, and effective failure avoidance. Additionally, we conduct simulation benchmarking on LIBERO \citep{liu2023libero}, SimplerEnv \citep{li2025evaluating} and RoboTwin 2.0 \citep{chen2025robotwin2}. The experimental results demonstrate that \da improves aggregate task success rates compared to base VLA policies. These findings suggest that test-time compute can be useful without being always-on; when it is invoked during critical phases, world model-based test-time dreaming can help correct local errors before failures occur, thereby enhancing the reliability of VLAs.

The contributions of our paper are summarized as follows:
(1) We emphasize the focus on critical phases and propose \da, a critical-phase test-time dreaming framework that helps VLAs foresee and avoid failures in advance. This framework keeps the base policy frozen and consists of a Dream Trigger, an Action Proposer, and a Dream Evaluator.
(2) We propose an autonomous boundary learning paradigm that endows the world models with failure awareness by introducing failure and boundary data.
(3) We demonstrate the advantages of explicit spatiotemporal dreaming in failure prediction through a pilot study. We conduct extensive evaluations on real-world manipulation tasks and simulation benchmarks. The results show that \da can effectively intervene before minor execution errors deteriorate into irrecoverable failures, improving overall task success rates.

\section{Pilot Study: Do We Need Dreaming to Avoid Failure in Critical Phases?} \label{sec:pilot_study}

\begin{table}[ht!]
\centering\scriptsize
\caption{Pilot study on failure prediction for fine-grained manipulation (Charger Plugging). All conditions use Gemini 3.6 Flash on the same 40 episodes. Brackets give Wilson 95\% confidence intervals for accuracy.}
\label{tab:pilot_study}
\begin{tabular}{lccc}
\toprule
Action representation & Accuracy (\%) & Failure precision & Failure recall\\
\midrule
Joint-space text & 57.5 [42.2, 71.5] & 56.4\% & 100.0\%\\
EEF-space text & 67.5 [52.0, 79.9] & 64.5\% & 90.9\%\\
DreamDojo video & 80.0 [65.2, 89.5] & 88.9\% & 72.7\%\\
\bottomrule
\end{tabular}
\end{table}

We first explore the paradigm of predicting fine-grained manipulation failures at test time. Given the strong semantic reasoning capabilities of VLMs, an intuitive baseline is to directly use them as consequence predictors during test time. We conducted a pilot study on the Charger Plugging task, which is highly sensitive to minor deviations. We constructed a test set containing 40 data episodes (18 successes and 22 failures), which were collected by rolling out a base policy $\pi_{0.5}$ \citep{pmlr-v305-black25a} trained on 100 human teleoperation demonstrations in the real world. To accurately evaluate the model's reasoning ability during the critical phase, we located a key interaction frame $t_{crit}$ within each test trajectory. This moment is defined as a frame within the 1-2 s immediately preceding the end-effector's decisive physical contact with the target object (e.g., just before the plug aligns with the socket). At each $t_{crit}$, we extract the current frame $o_t$ and a historical observation sequence. The action input is the downsampled future action chunk $u_t$.

We compare two test-time failure prediction paradigms:
(1) Direct Semantic Evaluation: We input the visual observation sequence containing the historical and current frames, along with the textified joint-space or end-effector (EEF)-space action sequence and its semantic description, into Gemini 3.6 Flash \citep{google2026gemini36flash}. We use few-shot prompting \citep{brown2020language} for it to predict end-to-end whether executing $u_t$ will result in a task failure (such as a collision or misalignment).
(2) Explicit Spatiotemporal Dreaming: We use DreamDojo \citep{gao2026dreamdojo}, an action-conditioned, large-scale pre-trained world model, to forward-render $u_t$ into future video frames, which are then evaluated by Gemini 3.6 Flash acting as a visual evaluator. Details are provided in Appendix \ref{app:pilot_study}.

The experimental results reveal the differences in prediction reliability between the two paradigms (Table \ref{tab:pilot_study}). The accuracy of the Direct Semantic Evaluation paradigm with joint-space text is only 57.5\%. Analysis shows that when faced with abstract numerical inputs, the VLM tends toward a conservative ``almost always predict failure'' pattern: it successfully identified all 22 failure samples, but also misclassified 17 successful samples as failures. Although the VLM can understand the semantic goal of ``Plug the charger into the socket,'' it struggles to establish an effective connection between the raw numerical action chunk and its actual physical consequences in this setting. In fine-grained manipulation, even a tiny deviation in the action trajectory can lead to catastrophic collisions. However, these minuscule numerical differences are challenging for the VLM to interpret reliably, limiting its ability to accurately predict the physical evolution of the trajectory.
Although the EEF-space text representation improves accuracy from 57.5\% to 67.5\%, notable limitations remain.

Conversely, Explicit Spatiotemporal Dreaming demonstrated stronger predictive capabilities, achieving an accuracy of 80.0\%. Although optimistic in some cases (missing 6 failures), it drastically suppressed false alarms (misclassified success cases dropped from 17 to 2). By explicitly simulating numerical actions forward into visual futures, the world model serves as a critical spatiotemporal bridge. It translates abstract numerical errors into explicit visual deviations, for example, directly dreaming of the plug hitting the edge of the socket or the tabletop. Subsequently, Gemini 3.6 Flash can assess this visual anomaly from the generated high-fidelity frames. This pilot study supports explicit spatiotemporal dreaming as a useful way to assess failure risks in critical phases.

\section{\da} \label{sec:method}

Based on the insights from the pilot study (\S \ref{sec:pilot_study}), we propose \da, a critical-phase test-time dreaming framework for VLA manipulation. The core idea is that the robot directly executes the frozen base VLA policy during most time steps, and only triggers test-time dreaming when it predicts an imminent transition into a critical phase. \da consists of three modules:
Dream Trigger, Action Proposer, and Dream Evaluator. 
\da allocates compute-intensive failure-aware reasoning to the truly important moments. Algorithm~\ref{alg:dreamavoid} summarizes execution.

\paragraph{Problem Formulation.}
We consider a trajectory $\tau = \{(o_t, l, a_t)\}_{t=1}^{T}$, where $o_t$ denotes the visual observation at time $t$, $l$ is the language instruction, and $a_t$ is the robot action. Parameterized via \textbf{flow matching}, the base VLA policy $\pi_\theta$ predicts an action chunk of horizon $H$, $u_t = [a_t, a_{t+1}, \dots, a_{t+H-1}]$, such that $u_t \sim \pi_\theta(\cdot \mid o_t, l)$.
Although directly executing the base policy is computationally efficient, it is susceptible to local errors during critical phases. Conversely, applying test-time compute or forward simulation at every replan can incur substantial additional computation. Therefore, our problem transforms into designing an adaptive inference strategy: the system identifies critical phases within $\tau$ to selectively allocate test-time compute.

\subsection{Dream Trigger}

\begin{figure*}[ht!]
  \centering
  \includegraphics[width=\linewidth]{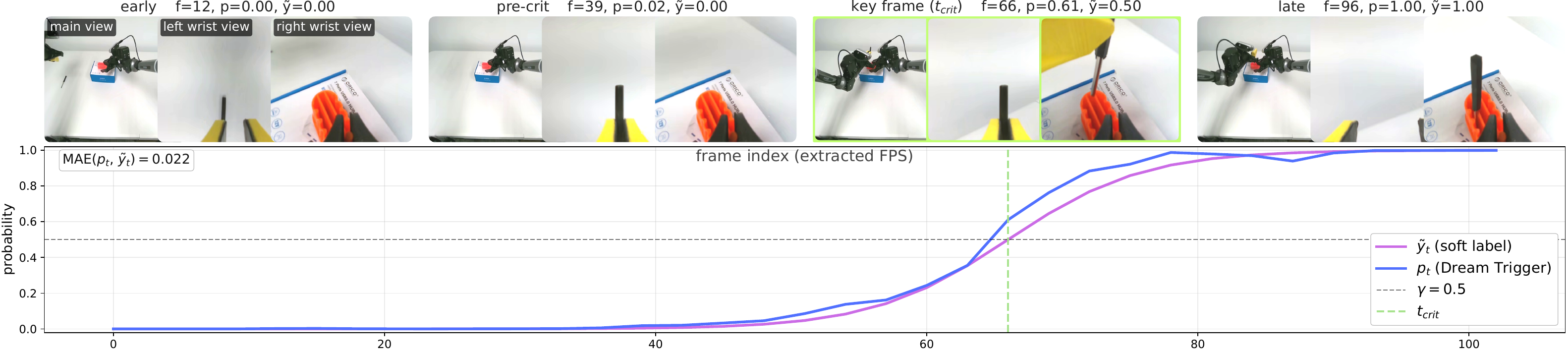}
  \caption{Qualitative example of the Dream Trigger ($\beta=5\ \text{frames}$ in the soft label $\tilde{y}_t$).}
  \label{fig:trigger_value}
\end{figure*}

The first step of \da is to determine whether the current state has entered a critical phase. We introduce a lightweight Dream Trigger $p_t=f_\phi(\mathcal{O}_{t-K+1:t},s_t)$, whose inputs are (1) single- or multi-view observations from the recent $K$ frames $\mathcal{O}_{t-K+1:t}=\{o_{t-K+1},\dots,o_t\}$, and (2) the current proprioceptive state $s_t$. Its output $p_t\in[0,1]$ is an onset-detection score. We implement the trigger with a frozen DINOv2 encoder and a three-layer MLP.
After collecting teleoperation demonstrations, we use Gemini 3.6 Flash offline to annotate two timestamps for each designated interaction: the critical-phase onset $t_{\mathrm{crit}}$ and the associated decisive-contact time $t_{\mathrm{contact}}$. The onset is identified within the 1--2\,s preceding this task-defined contact. We define the reference pre-contact intervention window as $\mathcal{I}=[t_{\mathrm{crit}},t_{\mathrm{contact}})$. The onset provides training supervision, while the interval provides an offline reference for evaluating detection timing.
Trigger training uses the hard target $y_t=\mathbb{I}[t\ge t_{\mathrm{crit}}]$. To smooth the transition around an uncertain annotated onset, we replace it with $\tilde{y}_t=\operatorname{Sigmoid}((t-t_{\mathrm{crit}})/\beta)$, where timestamps are expressed as video-frame indices and $\beta$ controls the transition scale. This target represents whether the onset has been passed and therefore remains high after contact. The deployment rule below controls whether a new intervention is issued. The Dream Trigger is supervised using binary cross-entropy with class-balanced weights:
$\mathcal{L}_{trigger} =-\frac{1}{T}\sum_{t=1}^{T} \left[ w_+\,\tilde{y}_t\log p_t +(1-\tilde{y}_t)\log(1-p_t) \right]. \label{eq:trigger_loss}$
The weight $w_+=N_-/N_+$ balances pre-onset and post-onset frames, where $N_-$ and $N_+$ are computed from the hard targets. For each real-world task, we use 100 training videos containing approximately 26,000--34,000 frames. Frame-level targets are constructed from the annotated onset of each supervised stage using the soft-label rule above.
At each policy decision when the evaluator is idle, we combine the trigger score with a persistent stage flag $d_t\in\{0,1\}$, initialized to zero at rollout reset. The intervention condition is $c_t^{\mathrm{test}}=(1-d_t)\mathbb{I}[p_t\ge\gamma]$. If $c_t^{\mathrm{test}}=0$, the system executes the base-policy action chunk. Otherwise, the scheduler sets $d_t=1$ and launches one cycle of action proposal, candidate evaluation, and selection. After executing the selected chunk, control returns to the same decision rule. The flag remains set until the rollout resets, preventing repeated calls for the same stage even if $p_t$ remains high. For multi-stage tasks, each designated stage has its own prediction head and flag, allowing later stages to trigger independently. The full scheduling rule is given in Appendix~\ref{app:trigger}.
We use $\beta=5$ frames and $\gamma=0.5$, with asynchronous trigger polling at 1\,Hz and serialized evaluator calls. Online decisions use causal visual observations, proprioception, and stage flags. Detector events and actual evaluator invocations are counted separately (Appendix~\ref{app:trigger_sensitivity}).

\subsection{Action Proposer}

When the system determines that it is currently in a critical phase, \da does not discard the base VLA policy, but instead continues to use it as a generator for candidate actions. In this paper, the base policy is modeled using flow matching, progressively mapping noise into an action chunk by integrating a conditional velocity field within the action space. During standard inference, this generation process corresponds to a deterministic probability flow ODE trajectory; thus, given an observation and language instruction, it typically outputs only one default action chunk. To obtain multiple candidate action chunks during testing, \da replaces this deterministic ODE inference process with its corresponding stochastic SDE sampling process \citep{liuflow,chen2025pirl}. Standard flow matching inference corresponds to the following probability flow ODE: $\mathrm{d}u = v_\theta(u, \rho \mid o_t, l)\,\mathrm{d}\rho$, where $\rho \in [0, 1]$ denotes the internal flow matching time variable, and $v_\theta$ is the conditional velocity field that determines the deterministic evolution trajectory of the action chunk from the initial noise distribution to the target distribution. To support multi-candidate sampling during testing, we add noise during integration to obtain the following stochastic process:
$\mathrm{d}u = v_\theta(u, \rho \mid o_t, l)\,\mathrm{d}\rho + \sigma(\rho)\,\mathrm{d}w_\rho$,
where $\sigma(\rho)$ controls the sampling noise intensity, and $w_\rho$ denotes the Wiener process. In practice, we adopt a constant noise scheduling strategy. The specific values of $\sigma$ for different tasks can be found in Table \ref{tab:hyperparameters}, Appendix \ref{app:action_proposer}. By repeating this stochastic sampling process, we can generate multiple candidate action chunks for the same time step to be used for the Dream Evaluator.

This design serves two primary purposes. First, all candidate actions still originate from the velocity field of the same pre-trained policy; therefore, they are not blind searches in an unconstrained action space, but are guided by the task semantics and action priors encoded in the base policy. Second, the randomness introduced by SDE allows the system to expose multiple locally plausible action hypotheses by perturbing the sampling trajectory beyond its initial noise. This is particularly crucial during critical phases: multiple seemingly reasonable action chunks might lead to drastically different downstream outcomes. For example, one candidate might accomplish precise alignment, while another might cause lateral deviation, collision, or object slipping.

\subsection{Dream Evaluator} \label{sec:dream_evaluator}

Action-conditioned world models provide a natural mechanism for evaluating the subsequent effects of candidate actions. DreamDojo \citep{gao2026dreamdojo}, which is pre-trained on large-scale videos and post-trained on robotics data, provides rich physical priors of manipulation tasks for test-time dreaming. Specifically, for each candidate action chunk $u_t^{(k)}$ and current observation $o_t$, we use the distilled DreamDojo (Appendix \ref{app:world_model}) to predict its corresponding future observation sequence $\hat{y}_t^{(k)} = \mathcal{W}_{\psi}(o_t, u_t^{(k)})$.
We define the value of a candidate action chunk as the estimated progress delta it brings within the short-horizon window. Built upon a VLM trained on a large-scale dataset, Robometer \citep{liang2026robometer} acts as a general-purpose robotic reward model capable of predicting frame-level task progress $\eta \in [0, 1]$. This allows us to automatically compute the offline progress pseudo-labels $z_t^{(k)}$ for the corresponding executed trajectories (Appendix \ref{app:value_model}). We use $z_t=\eta_{t+H}-\eta_t$, with $z_t=-1$ for labeled irreversible failures.
During the training of the value model, the model needs to learn accurate physical consequences from clear, real-world videos. However, during test time, it only receives video prediction $\hat{y}_t^{(k)}$ generated by DreamDojo, which may contain blurriness or artifacts. To balance learning efficiency and overcome this domain shift, we employ a \textit{real-dream joint training} strategy.

When constructing the inputs for the task-specific value model, (1) we directly input the \textit{real future sequence $y_{t}^{\text{true}}$} into the model (fitting $V_{\omega}(o_t, u_t, y_{t}^{\text{true}}) \rightarrow z_t$), enabling the network to learn precise value mappings based on observed visual features; (2) we input $(o_t, u_t)$ into DreamDojo to generate \textit{dreams $\hat{y}_t$} and use them as inputs as well (fitting $V_{\omega}(o_t, u_t, \hat{y}_t) \rightarrow z_t$). By treating the generated sequences as a form of in-domain data augmentation, we encourage the value model to learn to extract crucial physical progress features despite the visual imperfections of DreamDojo. During this process, the supervision targets $z_t$ required for the aforementioned fitting are derived from Robometer scores on the real future trajectories $y_{t}^{\text{true}}$ sampled from the offline dataset, with the irreversible-failure override described above.
Furthermore, the vast majority of routine offline teleoperation trajectories are in a state of steady progression (exhibiting small positive progress deltas). Direct random sampling would cause the model to lose its discriminative power for dangerous situations. To address this, DA-ABL uses priority-based oversampling when constructing the training batch $\mathcal{B}$: 20\% of the data is sampled from the terminal state horizons of successful completions, 40\% is drawn from the additionally collected boundary and failure data (see \S \ref{sec:co_improvement}, including labeled irreversible failures penalized to $z=-1$), and the remaining 40\% comes from routine steady progression phases.
We jointly use a Huber robust regression loss \citep{huber1992robust} and a continuous margin ranking loss:
\begin{equation}
\mathcal{L}_{\mathrm{VE}} ={} \frac{1}{|\mathcal{B}|}\sum_{i \in \mathcal{B}} L_{\delta}(\hat{v}^{(i)}, z^{(i)}) \nonumber +\frac{\lambda_{\mathrm{rank}}}{|\mathcal{P}|} \sum_{(i,j) \in \mathcal{P}}
\max\left(0, - (\hat{v}^{(i)} - \hat{v}^{(j)}) + \alpha (z^{(i)} - z^{(j)})\right),
\end{equation}
where $\hat{v}^{(i)}$ and $z^{(i)}$ are the predicted progress delta and the supervision target for the $i$-th flattened sample in batch $\mathcal{B}$, respectively. $\mathcal{P} = \{(i,j) \in \mathcal{B}\times\mathcal{B} \mid z^{(i)} > z^{(j)}\}$ is the set of valid training pairs, $L_{\delta}$ denotes the Huber loss, $\lambda_{\mathrm{rank}}$ balances the two loss terms, and $\alpha$ is a scaling factor requiring the network's predicted score difference to open up a corresponding margin. The ranking term is set to zero when no valid pairs exist. During testing, the Dream Evaluator selects the candidate action with the highest predicted progress delta $\hat{v}^{(k)}$.
At deployment, the distilled value model scores all eight candidates in approximately 40\,ms; Robometer is used offline to generate training labels.

\subsection{Dream the Boundaries: Autonomous Learning from Success, Failure \& Boundary Cases} \label{sec:co_improvement}

\begin{figure*}[ht!]
  \centering
  \includegraphics[width=\linewidth]{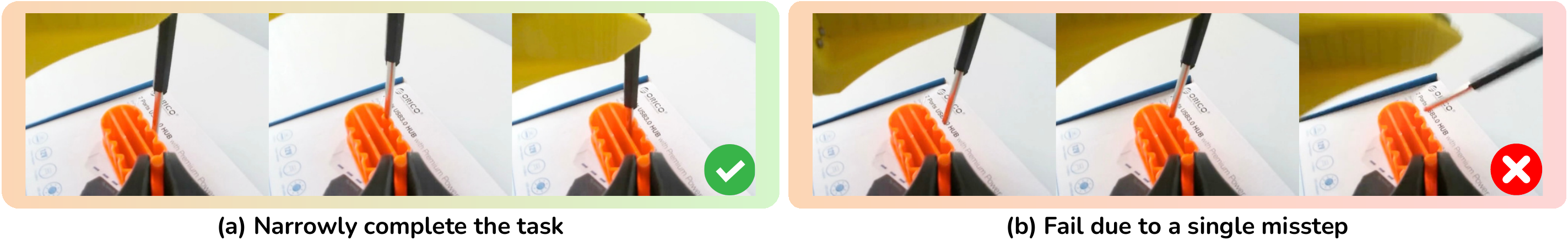}
  \caption{Boundary data in $\mathcal{D}_{online}$.}
  \label{fig:boundary}
\end{figure*}

\da introduces failure and boundary data to generate discriminative and informative ``dreams.'' Let $\mathcal{D}^+$ and $\mathcal{D}^-$ denote the sets of successful and failed trajectories, respectively. The base policy is trained exclusively on $\mathcal{D}^+$ and remains frozen during boundary learning and deployment. Its intuitive actions while ``awake'' are therefore based on successful demonstrations. Conversely, the Dream Evaluator, the engine for test-time dreaming, adapts its world and value models on $\mathcal{D}^+ \cup \mathcal{D}^-$.
The motivation behind this design is that, if trained only on successful data, the world model can generate ``optimistic dreams'' (consistent with the optimistic errors in \S \ref{sec:pilot_study} and related work \citep{guo2026vlaw,yin2026playworld}). It can learn what success looks like, but struggles to clearly identify which locally feasible actions will actually lead to failure. Especially in fine-grained manipulation tasks, visually subtle deviations can result in drastically different outcomes. By incorporating failure trajectories, the world and value models receive supervision on the critical boundaries between success and failure.
However, human teleoperation datasets $\mathcal{D}_{teleop}$ primarily consist of successful demonstrations, making it difficult to provide informative failure cases. To strengthen failure awareness in the world and value models, we propose an \textit{autonomous boundary learning} paradigm.
We deploy the initial version of \da (trained only on $\mathcal{D}_{teleop}$), obtaining a mixed dataset $\mathcal{D}_{online}$ that contains successful, boundary, and failed trajectories. At critical phases, the Action Proposer actively explores the action manifold and generates diverse candidate actions. Due to the optimistic hallucinations of the initial Dream Evaluator, it may misjudge locally plausible but globally dangerous actions as safe, thereby naturally exposing the system to trickier boundary cases. These boundary data can either result in narrowly completing the task or lead to failure by a single misstep (Figure \ref{fig:boundary}).
The collected empirical data is assigned value labels $z_t$ (defined in \S \ref{sec:dream_evaluator}). Subsequently, we combine $\mathcal{D}_{online}$ with $\mathcal{D}_{teleop}$ to fine-tune the Dream Evaluator, adapting its world model and training its value model.

\section{Real-World Experiments} \label{sec:real_world}

\begin{figure*}[ht!]
  \centering
  \includegraphics[width=0.9\linewidth]{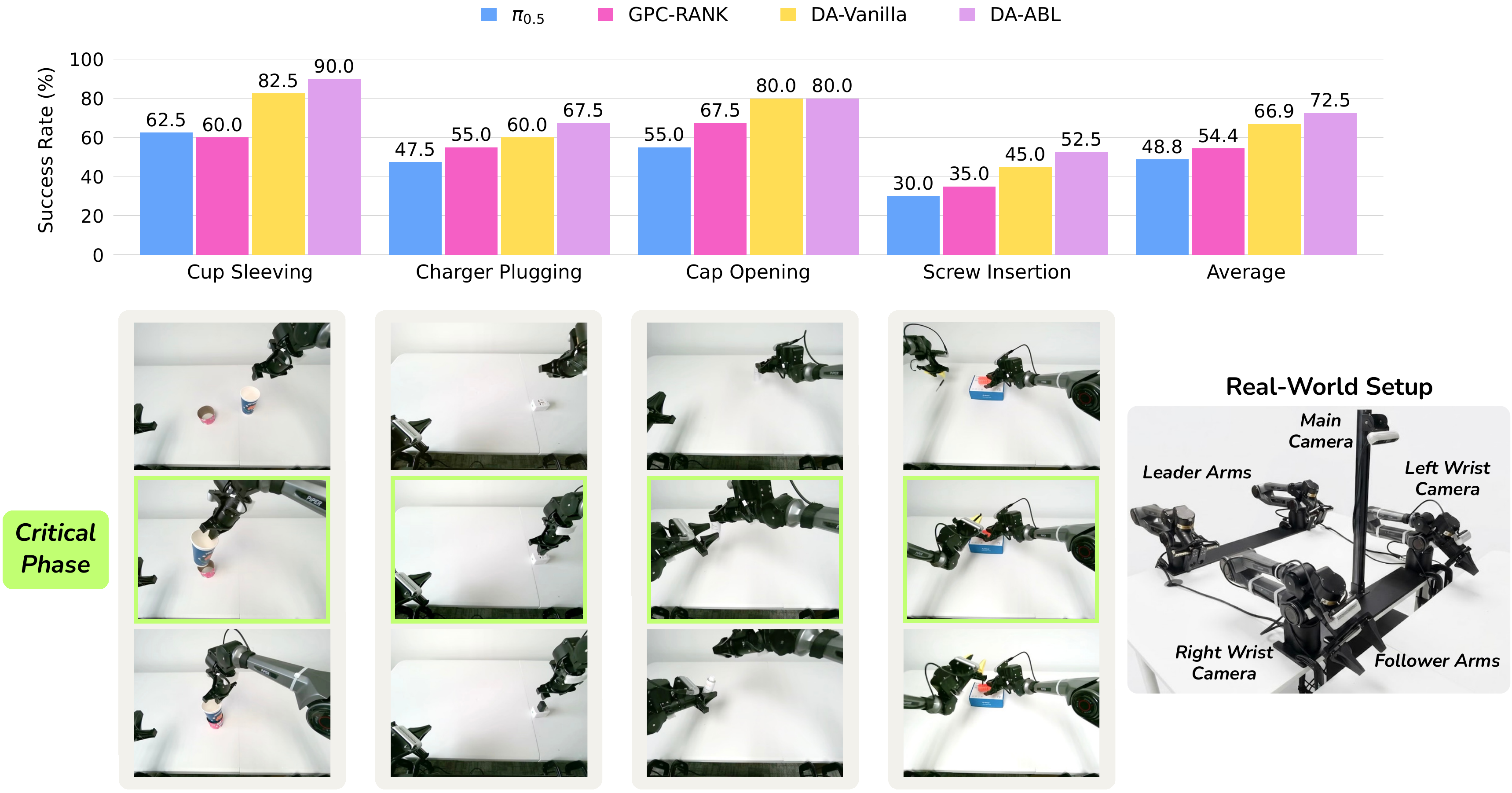}
  \caption{Comparisons in the real world. Each real-world task is evaluated over 40 independent trials.}
  \label{fig:result}
\end{figure*}

We conduct real-world experiments on the AgileX PiperX robotic platform, which is equipped with one front camera (RealSense D455) and two wrist cameras (RealSense D435), as shown in Figure \ref{fig:result}. The robot receives RGB observations and language instructions, and executes the action chunks output by the policy. We design four real-world manipulation tasks for evaluation that require accurate spatial alignment, stable contact establishment, and failure avoidance, including \textbf{Cup Sleeving}, \textbf{Charger Plugging}, \textbf{Cap Opening}, and \textbf{Screw Insertion}. Compared to typical scenarios, these tasks are much more sensitive to small action deviations. The specific task instructions and sample observations from the three viewpoints can be found in Table \ref{tab:task_info} and Appendix \ref{app:all_tasks_3view_sub}. Each method is evaluated over 40 independent trials per task (160 in total).

\paragraph{Evaluated Methods.}
To conduct a comprehensive evaluation, we compare \da with baseline methods and introduce different variants of \da.
We compare \da with the following baselines: (1) Base VLA policy: directly executes the action chunk output by $\pi_{0.5}$. (2) GPC-RANK \citep{qi2026inference}: this method repeatedly samples multiple candidate action chunks during testing and utilizes a world model for online ranking. We use GPC-RANK as the primary baseline to evaluate the combined design of \da: First, regarding the trigger timing, GPC-RANK is an always-on intervention, whereas \da implements critical-phase intervention. Second, in terms of consequence prediction, our Dream Evaluator combines physical priors from large-scale pre-training with adaptation on real failure and boundary data. Finally, \da{} uses a learned progress value to rank candidate actions. For consistency, both are evaluated based on the same base policy $\pi_{0.5}$.
To examine the contributions of our training paradigm (see \S \ref{sec:dream_evaluator} and \S \ref{sec:co_improvement}), we define two variants of our method based on the training data distribution: (1) \da-Vanilla (hereafter \textbf{DA-Vanilla}): Both the base VLA policy and the Dream Evaluator are trained exclusively on the human teleoperation dataset $\mathcal{D}_{teleop}$. (2) \da-ABL (hereafter \textbf{DA-ABL}): This represents the full \da framework incorporating Autonomous Boundary Learning. The base policy remains trained solely on $\mathcal{D}_{teleop}$, while the Dream Evaluator is fine-tuned on the mixed dataset $\mathcal{D}_{teleop} \cup \mathcal{D}_{online}$. The same base policy remains frozen in both variants.

\paragraph{Results.} \label{sec:real_result}
Figure~\ref{fig:result} summarizes the real-world results. \da consistently outperforms the base policy and the GPC-RANK method across all tasks. DA-ABL achieves the highest average success rate of 72.5\%, surpassing the base policy ($\pi_{0.5}$, 48.8\%) by 23.75 percentage points. Additionally, it surpasses the baseline method GPC-RANK (54.4\%) in average success rate by 18.125 percentage points. This demonstrates that selective, failure-aware test-time dreaming with rich physical priors is beneficial. Success counts, Wilson 95\% confidence intervals, and task-stratified CMH tests are reported in Appendix~\ref{app:statistics} (Table~\ref{tab:real_summary}).
Different variants of \da also allow us to examine the importance of our training paradigm. DA-Vanilla already yields an average improvement of 18.1 percentage points over the base policy. However, incorporating autonomous boundary learning (DA-ABL) brings an additional 5.6-point performance improvement. Since DA-Vanilla is trained exclusively on successful demonstration trajectories, its world model may generate optimistic dreams, which could limit its ability to intervene effectively during critical phases. In contrast, DA-ABL aims to enhance the Dream Evaluator's failure awareness by incorporating boundary and failure trajectories.

\da{} can reduce failures caused by lateral drift, premature collisions, and contact misalignment. In several experimental trajectories, the base policy produces reasonable actions at most time steps, but makes locally plausible yet globally detrimental decisions as it approaches the critical phase. By sampling multiple candidate action chunks and explicitly evaluating their short-horizon downstream consequences, \da is able to rank alternatives above these failure-prone proposals and instead execute actions that are more consistently aligned with task progression.

\paragraph{Deployment Cost.}
A key design of \da is to intervene only during the critical phase. This requires just 1--3 interventions per real-world trajectory. For the observed 300--480-frame rollouts, the estimated cumulative inference time is 2.7--7.3\,s for \da, compared with 8.3--13.9\,s for GPC-RANK under their respective deployment schedules. Detailed timing breakdowns and call-count estimation procedures are provided in Appendix~\ref{app:time_consumption}.

\paragraph{Ablation Study.} \label{sec:ablation}
We conduct ablation studies to analyze the contribution of the modules in \da to the performance improvement. All ablation experiments are conducted on real-world manipulation tasks.
\textbf{(1) Action Proposer vs. Repeated ODE.}
We analyze the impact of the candidate action generation method on system performance. To ensure a fair comparison, we construct a strong baseline (Repeated ODE + DA-ABL Dream Evaluator): it retains the Dream Evaluator of DA-ABL, but replaces the candidate action generation mechanism by inputting different random initial noises into the base policy and running deterministic flow inference multiple times to generate the same number of candidate actions. However, Table \ref{tab:ablation} shows that even when equipped with the same Dream Evaluator, the average success rate of Repeated ODE is only 57.5\%, which is below the full method DA-ABL using SDE sampling (72.5\%). This aligns with the principle of ``garbage in, garbage out'': if the Action Proposer cannot provide high-quality candidate solutions, leading to a severely homogenized candidate pool, then even if the Dream Evaluator can identify impending disasters, it cannot recover an action absent from the pool. This suggests that, compared to deterministic trajectories that only alter the noise at the initial timestep (ODE), the continuous random injection of SDE during the denoising process can explore the action manifold more effectively, potentially exposing richer and higher-quality feasible solutions. Such sample diversity can be important for avoiding subtle failure boundaries during critical phases.
\textbf{(2) Dream Evaluator vs. Random Select.}
We evaluate the contribution of world-model-based consequence evaluation by replacing the Dream Evaluator with random selection within the SDE candidate pool. Table \ref{tab:ablation} shows that the average success rate of the \textit{random select} variant is only 45.0\%, which is not only much lower than DA-ABL (72.5\%), but even lower than the base policy (48.8\%). This demonstrates that the candidate diversity created by the Action Proposer alone is not sufficient to avoid failures. In critical phases with extremely low error tolerance, blind diversity can instead introduce harmful jitters. Because disastrous physical conflicts may be hidden within locally reasonable candidate actions, once the system randomly selects these failure-prone actions, it can lead to irreversible collisions. Therefore, the system benefits from an evaluation mechanism capable of explicitly foreseeing the physical consequences of actions to pick out more promising action paths from diverse hypotheses.

\begin{table}[ht!]
\centering
\caption{Ablation study results.} 
\label{tab:ablation}
\scriptsize
\begin{tabular}{lccccccc}
\toprule
\multirow{2}{*}{Method} & \multicolumn{2}{c}{Components} & \multicolumn{4}{c}{Task Success Rate (\%)} & \multirow{2}{*}{Average} \\
\cmidrule(lr){2-3} \cmidrule(lr){4-7}
& Action Proposer & Dream Evaluator & Cup  & Charger  & Cap  & Screw  & \\
\midrule
$\pi_{0.5}$ & \texttimes (Single ODE) & \texttimes & 62.5 & 47.5 & 55.0 & 30.0 & 48.8 \\
\midrule
\multirow{2}{*}{\da Variants} & \texttimes (Repeated ODE) & \checkmark & \underline{67.5} & \underline{55.0} & \underline{70.0} & \underline{37.5} & \underline{57.5} \\
 & \checkmark (SDE) & \texttimes (Random Select) & 65.0 & 35.0 & 55.0 & 25.0 & 45.0 \\
\midrule
DA-ABL & \checkmark & \checkmark & \textbf{90.0} & \textbf{67.5} & \textbf{80.0} & \textbf{52.5} & \textbf{72.5} \\
\bottomrule
\end{tabular}
\end{table}

\section{RoboTwin, LIBERO, and SimplerEnv Benchmarks}
\label{sec:simulation}

We evaluate $\pi_{0.5}$ on RoboTwin 2.0 \citep{chen2025robotwin2} and LIBERO \citep{liu2023libero}, and GR00T-N1.6 on SimplerEnv \citep{li2025evaluating,bjorck2025gr00t}. Following the real-world procedure, we use Gemini 3.6 Flash solely offline to annotate critical-phase onsets using training videos and task instructions. These annotations train a lightweight Dream Trigger that schedules interventions using only causal visual observations and benchmark-provided proprioception at test time. Trigger and SDE sampling parameters are selected on a separate validation split and fixed before testing.
On RoboTwin 2.0, across eight bimanual tasks with 100 episodes each, DA-ABL increases success rate from 66.3\% to 70.8\%.
Across the four evaluated LIBERO settings, DA-ABL increases success from 96.5\% to 97.8\%, reducing failures from 28 to 18. On SimplerEnv, success increases from 59.9\% to 63.6\% on Bridge and from 76.4\% to 80.7\% on Fractal. These comparisons are conducted at the episode level and stratified by task. Full success rates and intervals are reported in Appendix~\ref{app:simulation}.

\section{Limitations and Future Work} \label{sec:limitation}

First, adapting the failure-aware framework \da to other policies (world action models or traditional control policies) is a promising direction. For real-world deployment, critical-phase labels are generated by Gemini 3.6 Flash, while trigger training, evaluator adaptation, and sampling settings remain task-specific.
Second, although \da strictly limits dense computations to sparse critical phases, generating high-fidelity future videos and scoring multiple candidate actions still introduces latency. Future work could not only employ various video generation acceleration techniques, but also opt to execute forward outcome simulations within a compact latent space.
Third, beyond optimizing decision-making at test-time, future work could leverage the high-fidelity videos generated by the world model as a rich source of training data, utilizing methods like advantage-weighted regression to continuously update the base policy itself.
A comprehensive discussion of limitations and future work is in Appendix \ref{app:limitation}.

\section{Conclusion}
This paper explores failure avoidance in VLA models. Although existing models possess strong intuitive execution capabilities, they still face limitations in anticipating risks. In contrast, explicit spatio-temporal simulation can provide useful foresight for failure avoidance. To this end, we propose the \da framework. Through critical-phase test-time dreaming, \da decouples intuitive execution from foresight by simulating the short-horizon futures of candidate actions during critical phases to select the highest-scoring candidate. To alleviate the challenge of state evaluation in complex scenarios, we introduce an autonomous boundary learning paradigm, enabling the model to better grasp the boundaries between success and failure. Experiments demonstrate that \da can help correct local errors before they escalate. In summary, we emphasize that test-time dreaming matters; integrating intuitive execution with selective dreaming at test time is an effective pathway to enhancing the reliability of VLAs. Through critical-phase dreaming, we can achieve more robust error avoidance capabilities while concentrating additional inference on detected critical phases.

\begingroup
\clearpage
\subsection*{AI Use Statement}
Gemini 3.6 Flash and Robometer are components of the research pipeline, as described in Sections~\ref{sec:pilot_study} and~\ref{sec:dream_evaluator}. Responsibility for the final text, claims, and empirical results rests with the authors.

\subsection*{Ethics Statement}
The reported improvements concern empirical task success and do not constitute certified collision avoidance or safe operation under distribution shift. Dreaming pauses and execution of the best available low-value candidate can still create physical risks. Real-world use therefore requires appropriate supervision and the robot's independent safety mechanisms.

\subsection*{Reproducibility Statement}
Appendices~\ref{app:pilot_study}--\ref{app:simulation} describe the observation/action setup, trigger training and sensitivity, candidate-sampling settings, evaluator targets, data sources, and evaluation counts. Appendix~\ref{app:statistics} specifies the confidence intervals and statistical tests. The anonymous code link is given in the abstract.
\par\endgroup

\bibliographystyle{iclr2027_conference}
\bibliography{reference}

\begin{thebibliography}{38}
\providecommand{\natexlab}[1]{#1}
\providecommand{\url}[1]{\texttt{#1}}
\expandafter\ifx\csname urlstyle\endcsname\relax
  \providecommand{\doi}[1]{doi: #1}\else
  \providecommand{\doi}{doi: \begingroup \urlstyle{rm}\Url}\fi

\bibitem[Bai et~al.(2025)Bai, Gao, and Shou]{bai2025evolve}
Zechen Bai, Chen Gao, and Mike~Zheng Shou.
\newblock Evolve-vla: Test-time training from environment feedback for vision-language-action models.
\newblock \emph{arXiv preprint arXiv:2512.14666}, 2025.

\bibitem[Bjorck et~al.(2025)Bjorck, Casta{\~n}eda, Cherniadev, Da, Ding, Fan, Fang, Fox, Hu, Huang, et~al.]{bjorck2025gr00t}
Johan Bjorck, Fernando Casta{\~n}eda, Nikita Cherniadev, Xingye Da, Runyu Ding, Linxi Fan, Yu~Fang, Dieter Fox, Fengyuan Hu, Spencer Huang, et~al.
\newblock Gr00t n1: An open foundation model for generalist humanoid robots.
\newblock \emph{arXiv preprint arXiv:2503.14734}, 2025.

\bibitem[Black et~al.(2025{\natexlab{a}})Black, Brown, Darpinian, Dhabalia, Driess, Esmail, Equi, Finn, Fusai, Galliker, Ghosh, Groom, Hausman, ichter, Jakubczak, Jones, Ke, LeBlanc, Levine, Li-Bell, Mothukuri, Nair, Pertsch, Ren, Shi, Smith, Springenberg, Stachowicz, Tanner, Vuong, Walke, Walling, Wang, Yu, and Zhilinsky]{pmlr-v305-black25a}
Kevin Black, Noah Brown, James Darpinian, Karan Dhabalia, Danny Driess, Adnan Esmail, Michael~Robert Equi, Chelsea Finn, Niccolo Fusai, Manuel~Y. Galliker, Dibya Ghosh, Lachy Groom, Karol Hausman, brian ichter, Szymon Jakubczak, Tim Jones, Liyiming Ke, Devin LeBlanc, Sergey Levine, Adrian Li-Bell, Mohith Mothukuri, Suraj Nair, Karl Pertsch, Allen~Z. Ren, Lucy~Xiaoyang Shi, Laura Smith, Jost~Tobias Springenberg, Kyle Stachowicz, James Tanner, Quan Vuong, Homer Walke, Anna Walling, Haohuan Wang, Lili Yu, and Ury Zhilinsky.
\newblock $\pi_{0.5}$: a vision-language-action model with open-world generalization.
\newblock In Joseph Lim, Shuran Song, and Hae-Won Park (eds.), \emph{Proceedings of The 9th Conference on Robot Learning}, volume 305, pp.\  17--40. PMLR, 2025{\natexlab{a}}.
\newblock URL \url{https://proceedings.mlr.press/v305/black25a.html}.

\bibitem[Black et~al.(2025{\natexlab{b}})Black, Brown, Driess, Esmail, Equi, Finn, Fusai, Groom, Hausman, Ichter, et~al.]{BlackK-RSS-25}
Kevin Black, Noah Brown, Danny Driess, Adnan Esmail, Michael Equi, Chelsea Finn, Niccolo Fusai, Lachy Groom, Karol Hausman, Brian Ichter, et~al.
\newblock $\pi_0$: A vision-language-action flow model for general robot control.
\newblock In \emph{Proceedings of Robotics: Science and Systems (RSS)}, 2025{\natexlab{b}}.

\bibitem[Brown et~al.(2020)Brown, Mann, Ryder, Subbiah, Kaplan, Dhariwal, Neelakantan, Shyam, Sastry, Askell, et~al.]{brown2020language}
Tom Brown, Benjamin Mann, Nick Ryder, Melanie Subbiah, Jared~D Kaplan, Prafulla Dhariwal, Arvind Neelakantan, Pranav Shyam, Girish Sastry, Amanda Askell, et~al.
\newblock Language models are few-shot learners.
\newblock \emph{Advances in neural information processing systems}, 33:\penalty0 1877--1901, 2020.

\bibitem[Chen et~al.(2025{\natexlab{a}})Chen, Liu, Zhang, Guo, Xu, Lin, Zang, Zhang, Yu, Fan, et~al.]{chen2025pirl}
Kang Chen, Zhihao Liu, Tonghe Zhang, Zhen Guo, Si~Xu, Hao Lin, Hongzhi Zang, Quanlu Zhang, Zhaofei Yu, Guoliang Fan, et~al.
\newblock $\pi$rl: Online rl fine-tuning for flow-based vision-language-action models.
\newblock \emph{arXiv preprint arXiv:2510.25889}, 2025{\natexlab{a}}.

\bibitem[Chen et~al.(2025{\natexlab{b}})Chen, Chen, Chen, Cai, Liu, Liang, Li, Lin, Ge, Gu, et~al.]{chen2025robotwin2}
Tianxing Chen, Zanxin Chen, Baijun Chen, Zijian Cai, Yibin Liu, Qiwei Liang, Zixuan Li, Xianliang Lin, Yiheng Ge, Zhenyu Gu, et~al.
\newblock Robotwin 2.0: A scalable data generator and benchmark with strong domain randomization for robust bimanual robotic manipulation.
\newblock \emph{arXiv preprint arXiv:2506.18088}, 2025{\natexlab{b}}.

\bibitem[Chi et~al.(2025)Chi, Xu, Feng, Cousineau, Du, Burchfiel, Tedrake, and Song]{chi2025diffusion}
Cheng Chi, Zhenjia Xu, Siyuan Feng, Eric Cousineau, Yilun Du, Benjamin Burchfiel, Russ Tedrake, and Shuran Song.
\newblock Diffusion policy: Visuomotor policy learning via action diffusion.
\newblock \emph{The International Journal of Robotics Research}, 44\penalty0 (10-11):\penalty0 1684--1704, 2025.

\bibitem[Duan et~al.(2024)Duan, Pumacay, Kumar, Wang, Tian, Yuan, Krishna, Fox, Mandlekar, and Guo]{duan2024aha}
Jiafei Duan, Wilbert Pumacay, Nishanth Kumar, Yi~Ru Wang, Shulin Tian, Wentao Yuan, Ranjay Krishna, Dieter Fox, Ajay Mandlekar, and Yijie Guo.
\newblock Aha: A vision-language-model for detecting and reasoning over failures in robotic manipulation.
\newblock \emph{arXiv preprint arXiv:2410.00371}, 2024.

\bibitem[Feng et~al.(2025)Feng, Han, Yang, Yue, Levine, and Luo]{feng2025reflective}
Yunhai Feng, Jiaming Han, Zhuoran Yang, Xiangyu Yue, Sergey Levine, and Jianlan Luo.
\newblock Reflective planning: Vision-language models for multi-stage long-horizon robotic manipulation.
\newblock In \emph{Conference on Robot Learning}, pp.\  2038--2062. PMLR, 2025.

\bibitem[Finn \& Levine(2017)Finn and Levine]{finn2017deep}
Chelsea Finn and Sergey Levine.
\newblock Deep visual foresight for planning robot motion.
\newblock In \emph{2017 IEEE international conference on robotics and automation (ICRA)}, pp.\  2786--2793. IEEE, 2017.

\bibitem[Gao et~al.(2026)Gao, Liang, Zheng, Malik, Ye, Yu, Tseng, Dong, Mo, Lin, et~al.]{gao2026dreamdojo}
Shenyuan Gao, William Liang, Kaiyuan Zheng, Ayaan Malik, Seonghyeon Ye, Sihyun Yu, Wei-Cheng Tseng, Yuzhu Dong, Kaichun Mo, Chen-Hsuan Lin, et~al.
\newblock Dreamdojo: A generalist robot world model from large-scale human videos.
\newblock \emph{arXiv preprint arXiv:2602.06949}, 2026.

\bibitem[{Google DeepMind}(2026)]{google2026gemini36flash}
{Google DeepMind}.
\newblock Gemini 3.6 flash model documentation.
\newblock \url{https://ai.google.dev/gemini-api/docs/models/gemini-3.6-flash}, 2026.

\bibitem[Gu et~al.(2025)Gu, Ju, Sun, Gilitschenski, Nishimura, Itkina, and Shkurti]{gu2025safe}
Qiao Gu, Yuanliang Ju, Shengxiang Sun, Igor Gilitschenski, Haruki Nishimura, Masha Itkina, and Florian Shkurti.
\newblock {SAFE}: Multitask failure detection for vision-language-action models.
\newblock In \emph{The Thirty-ninth Annual Conference on Neural Information Processing Systems}, 2025.
\newblock URL \url{https://openreview.net/forum?id=XPyAukgsFf}.

\bibitem[Guo et~al.(2025)Guo, Lu, Deng, Wu, Tang, and Wang]{guo2025vla}
Wenkai Guo, Guanxing Lu, Haoyuan Deng, Zhenyu Wu, Yansong Tang, and Ziwei Wang.
\newblock Vla-reasoner: Empowering vision-language-action models with reasoning via online monte carlo tree search.
\newblock \emph{arXiv preprint arXiv:2509.22643}, 2025.

\bibitem[Guo et~al.(2026)Guo, Lee, Shi, Chen, Liang, and Finn]{guo2026vlaw}
Yanjiang Guo, Tony Lee, Lucy~Xiaoyang Shi, Jianyu Chen, Percy Liang, and Chelsea Finn.
\newblock Vlaw: Iterative co-improvement of vision-language-action policy and world model.
\newblock \emph{arXiv preprint arXiv:2602.12063}, 2026.

\bibitem[Hu et~al.(2025)Hu, Liu, Liu, Cen, Meng, and He]{hu2025vlsa}
Songqiao Hu, Zeyi Liu, Shuang Liu, Jun Cen, Zihan Meng, and Xiao He.
\newblock Vlsa: Vision-language-action models with plug-and-play safety constraint layer.
\newblock \emph{arXiv preprint arXiv:2512.11891}, 2025.

\bibitem[Huang et~al.(2025)Huang, Li, He, Zhou, and Shechtman]{huang2025self}
Xun Huang, Zhengqi Li, Guande He, Mingyuan Zhou, and Eli Shechtman.
\newblock Self forcing: Bridging the train-test gap in autoregressive video diffusion.
\newblock \emph{arXiv preprint arXiv:2506.08009}, 2025.

\bibitem[Huber(1992)]{huber1992robust}
Peter~J Huber.
\newblock Robust estimation of a location parameter.
\newblock In \emph{Breakthroughs in statistics: Methodology and distribution}, pp.\  492--518. Springer, 1992.

\bibitem[Kim et~al.(2025)Kim, Pertsch, Karamcheti, Xiao, Balakrishna, Nair, Rafailov, Foster, Sanketi, Vuong, et~al.]{kim2025openvla}
Moo~Jin Kim, Karl Pertsch, Siddharth Karamcheti, Ted Xiao, Ashwin Balakrishna, Suraj Nair, Rafael Rafailov, Ethan~P Foster, Pannag~R Sanketi, Quan Vuong, et~al.
\newblock Openvla: An open-source vision-language-action model.
\newblock In \emph{Conference on Robot Learning}, pp.\  2679--2713. PMLR, 2025.

\bibitem[Kwok et~al.(2025)Kwok, Agia, Sinha, Foutter, Li, Stoica, Mirhoseini, and Pavone]{kwok2025robomonkey}
Jacky Kwok, Christopher Agia, Rohan Sinha, Matt Foutter, Shulu Li, Ion Stoica, Azalia Mirhoseini, and Marco Pavone.
\newblock Robomonkey: Scaling test-time sampling and verification for vision-language-action models.
\newblock In \emph{Conference on Robot Learning}, pp.\  3200--3217. PMLR, 2025.

\bibitem[Li et~al.(2026)Li, Lei, Zang, Hu, Liang, An, Li, and Xu]{li2026failure}
Huanyu Li, Kun Lei, Sheng Zang, Kaizhe Hu, Yongyuan Liang, Bo~An, Xiaoli Li, and Huazhe Xu.
\newblock Failure-aware rl: Reliable offline-to-online reinforcement learning with self-recovery for real-world manipulation.
\newblock \emph{arXiv preprint arXiv:2601.07821}, 2026.

\bibitem[Li et~al.(2025)Li, Hsu, Gu, Mees, Pertsch, Walke, Fu, Lunawat, Sieh, Kirmani, et~al.]{li2025evaluating}
Xuanlin Li, Kyle Hsu, Jiayuan Gu, Oier Mees, Karl Pertsch, Homer~Rich Walke, Chuyuan Fu, Ishikaa Lunawat, Isabel Sieh, Sean Kirmani, et~al.
\newblock Evaluating real-world robot manipulation policies in simulation.
\newblock In \emph{Conference on Robot Learning}, pp.\  3705--3728. PMLR, 2025.

\bibitem[Liang et~al.(2026)Liang, Korkmaz, Zhang, Hwang, Anwar, Kaushik, Shah, Huang, Zettlemoyer, Fox, et~al.]{liang2026robometer}
Anthony Liang, Yigit Korkmaz, Jiahui Zhang, Minyoung Hwang, Abrar Anwar, Sidhant Kaushik, Aditya Shah, Alex~S Huang, Luke Zettlemoyer, Dieter Fox, et~al.
\newblock Robometer: Scaling general-purpose robotic reward models via trajectory comparisons.
\newblock \emph{arXiv preprint arXiv:2603.02115}, 2026.

\bibitem[Liu et~al.(2023{\natexlab{a}})Liu, Zhu, Gao, Feng, Liu, Zhu, and Stone]{liu2023libero}
Bo~Liu, Yifeng Zhu, Chongkai Gao, Yihao Feng, Qiang Liu, Yuke Zhu, and Peter Stone.
\newblock Libero: Benchmarking knowledge transfer for lifelong robot learning.
\newblock \emph{Advances in Neural Information Processing Systems}, 36:\penalty0 44776--44791, 2023{\natexlab{a}}.

\bibitem[Liu et~al.(2026)Liu, Liu, Wang, Zhuang, Liang, Yang, Xu, Wang, Liu, and Han]{liu2026fly}
Changyu Liu, Yiyang Liu, Taowen Wang, Qiao Zhuang, James~Chenhao Liang, Wenhao Yang, Renjing Xu, Qifan Wang, Dongfang Liu, and Cheng Han.
\newblock On-the-fly vla adaptation via test-time reinforcement learning.
\newblock \emph{arXiv preprint arXiv:2601.06748}, 2026.

\bibitem[Liu et~al.(2025)Liu, Liu, Liang, Li, Liu, Wang, Wan, ZHANG, and Ouyang]{liuflow}
Jie Liu, Gongye Liu, Jiajun Liang, Yangguang Li, Jiaheng Liu, Xintao Wang, Pengfei Wan, Di~ZHANG, and Wanli Ouyang.
\newblock Flow-grpo: Training flow matching models via online rl.
\newblock In \emph{The Thirty-ninth Annual Conference on Neural Information Processing Systems}, 2025.

\bibitem[Liu et~al.(2023{\natexlab{b}})Liu, Bahety, and Song]{liu2023reflect}
Zeyi Liu, Arpit Bahety, and Shuran Song.
\newblock Reflect: Summarizing robot experiences for failure explanation and correction.
\newblock In \emph{Conference on Robot Learning}, pp.\  3468--3484. PMLR, 2023{\natexlab{b}}.

\bibitem[Onishi et~al.(2026)Onishi, Takizawa, Ohmura, and Kuniyoshi]{onishi2026exploration}
Yujiro Onishi, Ryo Takizawa, Yoshiyuki Ohmura, and Yasuo Kuniyoshi.
\newblock Exploration-assisted bottleneck transition toward robust and data-efficient deformable object manipulation.
\newblock \emph{arXiv preprint arXiv:2603.13756}, 2026.

\bibitem[Oquab et~al.(2023)Oquab, Darcet, Moutakanni, Vo, Szafraniec, Khalidov, Fernandez, Haziza, Massa, El-Nouby, et~al.]{oquab2023dinov2}
Maxime Oquab, Timoth{\'e}e Darcet, Th{\'e}o Moutakanni, Huy Vo, Marc Szafraniec, Vasil Khalidov, Pierre Fernandez, Daniel Haziza, Francisco Massa, Alaaeldin El-Nouby, et~al.
\newblock Dinov2: Learning robust visual features without supervision.
\newblock \emph{arXiv preprint arXiv:2304.07193}, 2023.

\bibitem[Qi et~al.(2026)Qi, Yin, Zhu, Du, and Yang]{qi2026inference}
Han Qi, Haocheng Yin, Aris Zhu, Yilun Du, and Heng Yang.
\newblock Inference-time enhancement of generative robot policies via predictive world modeling.
\newblock \emph{IEEE Robotics and Automation Letters}, 2026.

\bibitem[Tur et~al.(2026)Tur, Naghiyev, Fang, Tsai, Duan, Fox, and Krishna]{tur2026recurrentdepth}
Yalcin Tur, Jalal Naghiyev, Haoquan Fang, Wei-Chuan Tsai, Jiafei Duan, Dieter Fox, and Ranjay Krishna.
\newblock {RECURRENT}-{DEPTH} {VLA}: {IMPLICIT} {TEST}-{TIME} {COMPUTE} {SCALING} {OF} {VISION}{\textendash}{LANGUAGE}{\textendash}{ACTION} {MODELS} {VIA} {LATENT} {ITERATIVE} {REASONING}.
\newblock In \emph{Workshop on Latent {\&} Implicit Thinking {\textendash} Going Beyond CoT Reasoning}, 2026.
\newblock URL \url{https://openreview.net/forum?id=hsIm52gD9p}.

\bibitem[Wu et~al.(2025)Wu, Liu, Zhao, Qiu, Li, Che, Xu, and Tang]{wu2025learning}
Kun Wu, Ning Liu, Zhen Zhao, Di~Qiu, Jinming Li, Zhengping Che, Zhiyuan Xu, and Jian Tang.
\newblock Learning from imperfect demonstrations with self-supervision for robotic manipulation.
\newblock In \emph{2025 IEEE International Conference on Robotics and Automation (ICRA)}, pp.\  16899--16906. IEEE, 2025.

\bibitem[Xu et~al.(2026)Xu, Springenberg, Equi, Amin, Esmail, Levine, and Ke]{xu2026rl}
Charles Xu, Jost~Tobias Springenberg, Michael Equi, Ali Amin, Adnan Esmail, Sergey Levine, and Liyiming Ke.
\newblock Rl token: Bootstrapping online rl with vision-language-action models.
\newblock \emph{arXiv preprint arXiv:2604.23073}, 2026.

\bibitem[Yang et~al.(2026)Yang, Gao, Bu, Chen, and Metaxas]{yang2026seeing}
Yanting Yang, Shenyuan Gao, Qingwen Bu, Li~Chen, and Dimitris~N Metaxas.
\newblock Seeing farther and smarter: Value-guided multi-path reflection for vlm policy optimization.
\newblock \emph{arXiv preprint arXiv:2602.19372}, 2026.

\bibitem[Ye et~al.(2026)Ye, Ge, Zheng, Gao, Yu, Kurian, Indupuru, Tan, Zhu, Xiang, et~al.]{ye2026world}
Seonghyeon Ye, Yunhao Ge, Kaiyuan Zheng, Shenyuan Gao, Sihyun Yu, George Kurian, Suneel Indupuru, You~Liang Tan, Chuning Zhu, Jiannan Xiang, et~al.
\newblock World action models are zero-shot policies.
\newblock In \emph{ICLR 2026 the 2nd Workshop on World Models: Understanding, Modelling and Scaling}, 2026.

\bibitem[Yin et~al.(2026)Yin, Mei, Zheng, Yamane, Wang, Sceats, Bateman, Zha, Badithela, Shorinwa, et~al.]{yin2026playworld}
Tenny Yin, Zhiting Mei, Zhonghe Zheng, Miyu Yamane, David Wang, Jade Sceats, Samuel~M Bateman, Lihan Zha, Apurva Badithela, Ola Shorinwa, et~al.
\newblock Playworld: Learning robot world models from autonomous play.
\newblock \emph{arXiv preprint arXiv:2603.09030}, 2026.

\bibitem[Zhang et~al.(2025)Zhang, Dai, Solak, Zhou, She, and Ajoudani]{zhang2025safe}
Heng Zhang, Rui Dai, Gokhan Solak, Pokuang Zhou, Yu~She, and Arash Ajoudani.
\newblock Safe learning for contact-rich robot tasks: A survey from classical learning-based methods to safe foundation models.
\newblock \emph{arXiv preprint arXiv:2512.11908}, 2025.

\end{thebibliography}
\clearpage
\appendix

\appendix

\section{Related Work}

\paragraph{Brittle VLA Fine-Grained Manipulation.}

VLA models combine large-scale vision-language pretraining with robotic action generation, substantially improving robots’ semantic understanding and cross-task generalization abilities \citep{kim2025openvla,pmlr-v305-black25a,BlackK-RSS-25}. However, recent studies suggest that the bottleneck of VLA models often lies not in insufficient global semantic understanding, but in local decision errors at a few highly sensitive moments \citep{xu2026rl}. As a result, VLA models remain brittle in tasks that require fine-grained alignment, contact control, and obstacle avoidance \citep{zhang2025safe,hu2025vlsa}. Building on this observation, this paper treats failures in fine-grained manipulation as a temporally localized critical phase problem.

\paragraph{Test-Time Compute.}

To enhance deployment robustness, existing works have begun introducing additional computation into the testing phase. The differences between our method and existing test-time compute mainly lie in two aspects.
(1) In terms of trigger timing, existing works tend to favor always-on computation \citep{liu2026fly,tur2026recurrentdepth,qi2026inference,bai2025evolve,feng2025reflective}. TT-VLA \citep{liu2026fly} and EVOLVE-VLA \citep{bai2025evolve} utilize dense progress rewards for constant online policy adaptation during inference. RD-VLA achieves adaptive computation depth through latent recurrent inference \citep{tur2026recurrentdepth}. GPC invokes its world model to evaluate candidate actions at every replanning step \citep{qi2026inference}.
In contrast, our method \da directly executes the base VLA policy most of the time, triggering additional computation only when the Dream Trigger predicts entry into a temporally sparse yet consequence-sensitive critical phase.
(2) In terms of evaluation methods, most Verifier or MPC works predominantly rely on current-state scoring or pixel-level goal alignment \citep{finn2017deep,kwok2025robomonkey}. Visual MPC compares candidate futures based on action-conditioned video prediction, but its scoring criterion is the probability of designated pixels reaching predefined target positions \citep{finn2017deep}. RoboMonkey investigates VLA test-time scaling by Gaussian perturbation sampling and selecting from candidate actions using a VLM-based verifier \citep{kwok2025robomonkey}.
GPC combines Diffusion Policy \citep{chi2025diffusion} with a world model to evaluate candidate actions during deployment, but it does not leverage the rich physical priors embedded in large-scale pre-trained world models \citep{qi2026inference}. In comparison, we utilize the high-fidelity future simulation capability of large-scale pre-trained world models.
\da couples critical-phase-triggered, failure-aware short-horizon dreaming with autonomous boundary learning. By collecting success, failure, and boundary data through deployment, we refine the world model's predictions near success--failure transitions to better support preventive action selection.

\paragraph{Failure Detection.}
Handling failures is typically viewed as a reactive process: the system must first detect the occurrence of an error and subsequently attempt to correct it. For instance, AHA \cite{duan2024aha} introduces an instruction-fine-tuned VLM specifically designed to detect and reason about manipulation failures using natural language. REFLECT \cite{liu2023reflect} explains failures and proposes corrective suggestions by summarizing the robot's experiences. Furthermore, SAFE proposes a multi-task failure detector for general-purpose VLAs, which predicts the probability of task failure in real-time by analyzing the model's internal latent features \cite{gu2025safe}. While these post-hoc reasoning capabilities are invaluable for offline policy improvement, reward generation, and sub-task verification, they inherently operate only after a failure has already occurred.


To achieve a safer execution process, our work shifts the reasoning focus of the system from reactive failure detection to proactive avoidance. Instead of waiting for a failure to happen before reasoning, \da utilizes test-time dreaming to explicitly forward-simulate the consequences of candidate actions. By invoking this failure-aware evaluation at critical phases, \da is able to intercept minor action deviations before they accumulate into irreversible failures. Consequently, our approach acts as a preventive safety layer, effectively complementing the aforementioned works.

\paragraph{World Models.} 
World models provide a natural mechanism for evaluating the downstream consequences of candidate actions. DreamDojo shows that action-conditioned world models, pretrained on large-scale video data and then post-trained on robot data, can support physically meaningful future prediction \citep{gao2026dreamdojo}. Prior work such as VLAW \citep{guo2026vlaw} and FARL \citep{li2026failure} show that failed trajectories are important for improving robot robustness, recovery ability, and feasibility judgment.

\section{Pilot Study Details} \label{app:pilot_study}

To evaluate the model's reasoning capabilities during the critical phase of fine-grained manipulation, we establish a standardized evaluation setup.

\paragraph{Common Setup.} 
Gemini 3.6 Flash identified a critical interaction frame $t_{crit}$ in each test trajectory offline from the recorded video and task instruction. At each $t_{crit}$, we extract the current frame $o_t$ along with a historical observation sequence consisting of 5 frames with a sampling stride of 8. Each frame is constructed by concatenating three camera views (the main view and the left and right wrist views) into a 2×2 grid (with the bottom-right corner filled in black).

At this moment, the base policy $\pi_{0.5}$ predicts an action chunk of length $H=50$ (approximately 1.67 s in a 30 fps video). The action consists of 6-dimensional absolute joint angles and the gripper state for both the left and right arms, totaling 14 dimensions. To align with the inference frequency of the world model and reduce the text comprehension burden on the VLM, we downsample the future action chunk using a stride of 4, extracting 13 actions with indices $[0, 4, 8, \dots, 48]$. This sparse action sequence $u_t$ spans 48 time steps, corresponding to an execution window of approximately 1.6 s into the future.

\begin{figure*}[ht!]
    \centering
    \begin{subfigure}{0.85\textwidth}
        \centering
        \includegraphics[width=\textwidth]{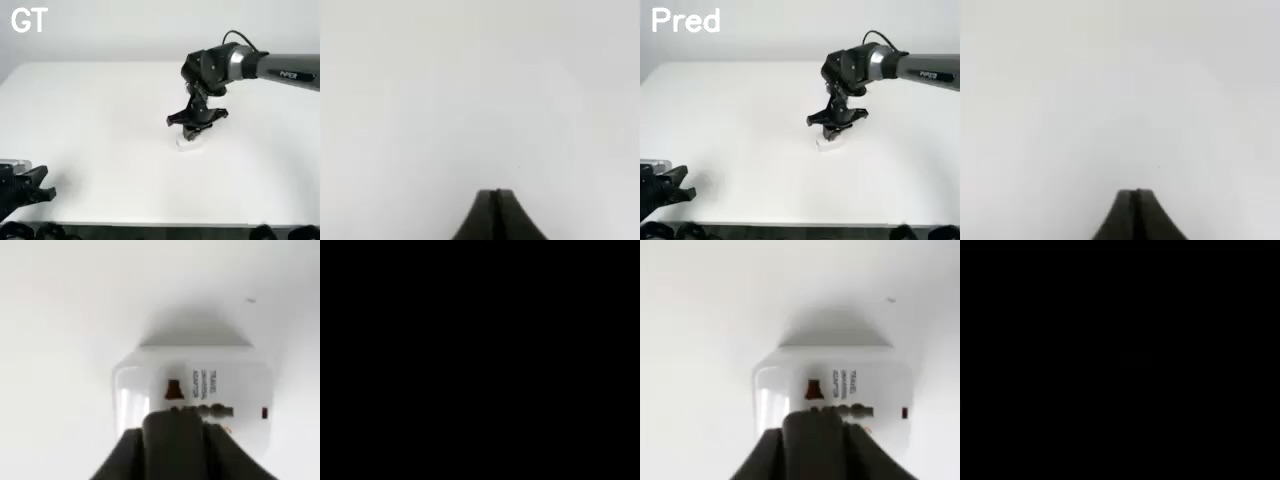}
        \caption{}
    \end{subfigure}
    
    \vspace{0.5em}
    
    \begin{subfigure}{0.85\textwidth}
        \centering
        \includegraphics[width=\textwidth]{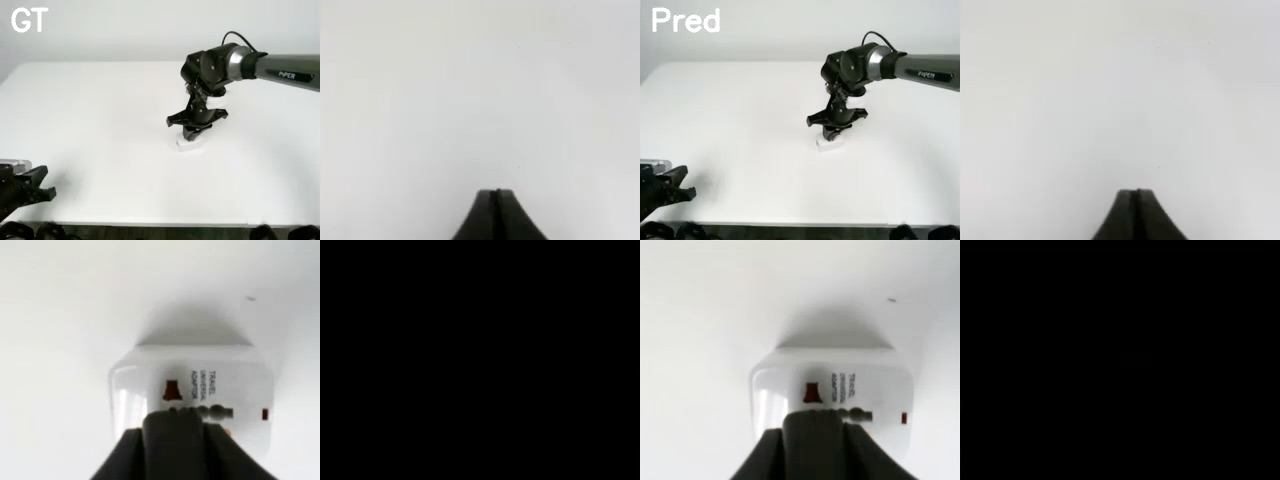}
        \caption{}
    \end{subfigure}
    
    \vspace{0.5em}
    
    \begin{subfigure}{0.85\textwidth}
        \centering
        \includegraphics[width=\textwidth]{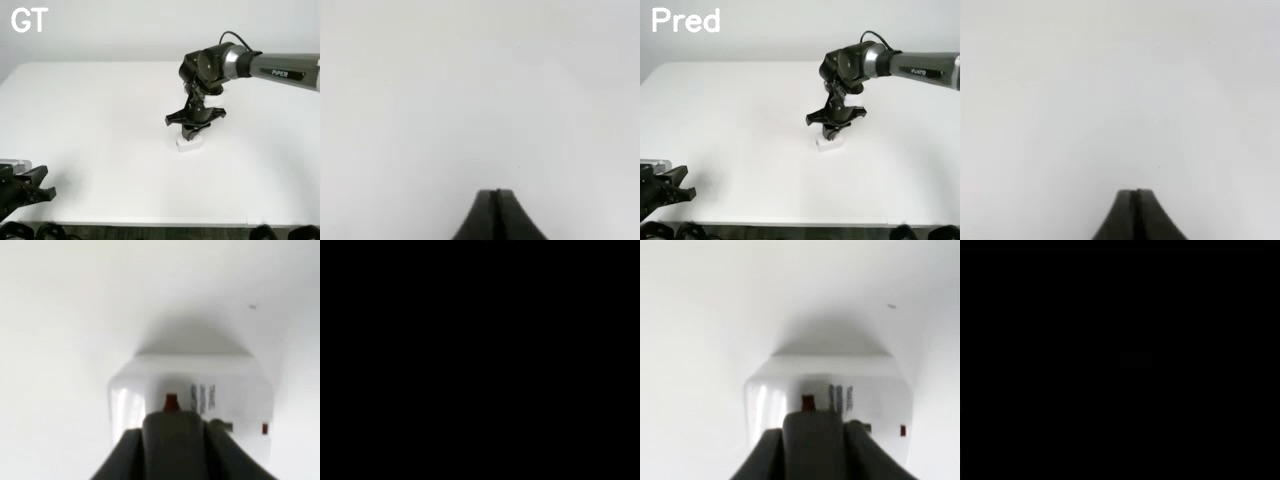}
        \caption{}
    \end{subfigure}
    \caption{(a) shows the current frame $t_{crit}$, while (b) and (c) show the actual future outcome frames (GT) and the video frames predicted by DreamDojo (Pred). Each 2x2 image block contains different camera views: the top-left is the main view, the top-right is the left wrist view, and the bottom-left is the right wrist view. This concatenation method follows the approach of DreamZero \citep{ye2026world} to accommodate the single-image input requirement of DreamDojo.} 
    \label{fig:pilot_study}
\end{figure*}

\paragraph{Setup for Explicit Spatiotemporal Dreaming.} 
The historical observation stride is 8, and the future action sequence stride is 4. A sparser visual history is sufficient to capture the macroscopic motion context. However, accurately predicting the physical evolution of the critical phase, which is sensitive to minute deviations, requires denser action-conditioned inputs to ensure that the world model can perform precise spatiotemporal rendering. The world model DreamDojo is fine-tuned on 100 human teleoperation trajectories. Figure \ref{fig:pilot_study} illustrates the actual execution alongside the future frames predicted by the world model.

\section{\da Framework Details} \label{app:method}

The complete execution procedure of \da is summarized in Algorithm \ref{alg:dreamavoid}.

\begin{algorithm}[htbp]

\caption{\da}\label{alg:dreamavoid}
\scriptsize
\begin{tabularx}{\linewidth}{rX}
1 & Observe current state $o_t$ and instruction $l$.\\
2 & Compute criticality $p_t=f_\phi(\mathcal O_{t-K+1:t},s_t)$ (Dream Trigger).\\
3 & \textbf{if} $p_t<\gamma$ \textbf{then}\\
4 & \quad Execute default chunk $u_t\sim\pi_\theta(\cdot\mid o_t,l)$.\\
5 & \textbf{else}\\
6 & \quad Sample $\{u_t^{(k)}\}_{k=1}^{N_c}$ via the stochastic flow solver (Action Proposer).\\
7 & \quad \textbf{for} $k=1$ to $N_c$ \textbf{do}\\
8 & \qquad Predict $\hat y_t^{(k)}=\mathcal W_\psi(o_t,u_t^{(k)})$.\\
9 & \qquad Score $\hat v_t^{(k)}=V_\omega(o_t,u_t^{(k)},\hat y_t^{(k)})$.\\
10 & \quad \textbf{end for}\\
11 & \quad Select $k^*=\arg\max_k\hat v_t^{(k)}$.\\
12 & \quad Execute $u_t^{(k^*)}$.\\
13 & \textbf{end if}\\
\end{tabularx}
\end{algorithm}

\begingroup
\noindent The argmax in Algorithm~\ref{alg:dreamavoid} is the complete action-selection rule. The selected action is executed even when every candidate falls below it. 
\par\endgroup

\subsection{Dream Trigger} \label{app:trigger}

\begingroup
Critical-phase onsets for trigger training are annotated offline by Gemini 3.6 Flash from task videos and instructions. Approximately 26,000--34,000 frame labels were obtained from 100 training videos per task. Gemini 3.6 Flash is used only for offline annotation; the learned trigger uses causal visual observations and proprioception at test time.
\par\endgroup

Considering the requirements for real-time inference speed, we select the frozen DINOv2-ViT-B/14 \citep{oquab2023dinov2} as the backbone for Dream Trigger.
First, we perform frame-by-frame feature encoding on the images from each camera feed (which include sparsely sampled historical observations and the current frame) and execute average pooling along the temporal dimension. Subsequently, the aggregated single- or multi-camera visual features are concatenated with the current proprioceptive state $s_t$ and jointly fed into a three-layer MLP classification head to determine the current phase state.
In real-world deployment, we configure Dream Trigger to poll asynchronously at a frequency of 1 Hz, which is shorter than the 1.67 s execution cycle of the action chunk.

\begingroup
On a single RTX 4090, trigger inference takes approximately 50\,ms per poll. It runs in a separate process alongside policy inference, with reported average GPU utilization below 5\% and no observable additional control-loop delay in this setup.
\par\endgroup


\begingroup
For a task with $M>1$ designated interaction stages (e.g., grasping → alignment), the trigger outputs $\boldsymbol{p}_t=[p_t^{(1)},\ldots,p_t^{(M)}]$. Each head corresponds to a task-defined stage with its own annotated onset $t_{\mathrm{crit}}^{(m)}$, contact time $t_{\mathrm{contact}}^{(m)}$, and reference window $\mathcal{I}^{(m)}=[t_{\mathrm{crit}}^{(m)},t_{\mathrm{contact}}^{(m)})$. Its training target is $\tilde{y}_t^{(m)} = \operatorname{Sigmoid}((t-t_{\mathrm{crit}}^{(m)})/\beta)$. The heads share the visual backbone and MLP, with an $M$-dimensional final layer, and are trained by summing their binary cross-entropy losses:
\begin{equation}
    \mathcal{L}_{\mathrm{trigger}}^{\mathrm{multi}}
    =\sum_{m=1}^{M}\mathcal{L}_{\mathrm{trigger}}^{(m)}.
\end{equation}

Each stage has a persistent flag $d_t^{(m)}$, initialized to zero at rollout reset. At an available policy decision, the eligible stages are
\begin{equation}
    \mathcal{A}_t
    =\left\{m:
    d_t^{(m)}=0,\;
    p_t^{(m)}\ge\gamma
    \right\},
    \qquad
    c_t^{\mathrm{test}}
    =\mathbb{I}\!\left[\mathcal{A}_t\neq\varnothing\right].
\end{equation}
If $\mathcal{A}_t$ is nonempty, the scheduler launches one evaluation cycle and sets $d_t^{(m)}=1$ for all $m\in\mathcal{A}_t$. Simultaneously eligible stages share this invocation. Other stage flags remain unchanged. Thus, a previously triggered head cannot generate repeated calls even if its score remains high, while an untriggered stage can initiate a later intervention. All flags are cleared only when the rollout resets. This rule permits at most one invocation per designated stage and at most $M$ invocations per rollout.
\par\endgroup

\begingroup
\subsubsection{Held-Out Detection and Sensitivity} \label{app:trigger_sensitivity}
We evaluate the trigger on 80 held-out videos across the four real-world tasks, with critical-phase reference annotations generated offline by Gemini 3.6 Flash. Consecutive positive polls are merged into one intervention event. An event is correct if it occurs within the model-annotated pre-contact intervention window; recall counts whether the window contains at least one detection. Event precision uses detected events as its denominator, whereas recall uses the 80 annotated phases. 

\begin{table}[ht!]

\centering\scriptsize
\caption{Trigger threshold sensitivity on 80 held-out videos. Intervention counts merge consecutive positive polls. Only the three reported thresholds are shown.}
\label{tab:gamma_sensitivity}
\begin{tabular}{cccc}
\toprule
$\gamma$ & Event precision & Phase recall & Events / trajectory\\
\midrule
0.3 & 84.9\% (79/93) & 98.8\% (79/80) & 1.16\\
0.5 & 92.7\% (76/82) & 95.0\% (76/80) & 1.03\\
0.7 & 97.2\% (70/72) & 87.5\% (70/80) & 0.90\\
\bottomrule
\end{tabular}
\end{table}

\begin{table}[ht!]

\centering\scriptsize
\caption{Soft-label sensitivity after retraining the Charger Plugging trigger on the same training set.}
\label{tab:beta_sensitivity}
\begin{tabular}{cccc}
\toprule
$\beta$ (frames) & Event precision & Phase recall & Events / trajectory\\
\midrule
2 & 94.7\% & 90.0\% & 0.95\\
5 & 95.0\% & 95.0\% & 1.00\\
10 & 87.0\% & 100.0\% & 1.15\\
\bottomrule
\end{tabular}
\end{table}

A lower $\gamma$ increases coverage and intervention count, while a higher value misses more phases. At the default $\gamma=0.5$, event precision is 92.7\% (76/82), phase recall is 95.0\% (76/80), and the missed-phase rate is 5.0\% (4/80). A narrower $\beta$ loses recall on Charger Plugging; a broader transition increases activations.
\par\endgroup

\subsection{Action Proposer} \label{app:action_proposer}

\begingroup
Table~\ref{tab:hyperparameters} lists the noise level $\sigma$, candidate count $N_c$, and action horizon $H$. Both \da variants share these settings. GPC-RANK uses its own sampling mechanism, with the same $N_c$ and $H$ for each environment. LIBERO noise settings follow $\pi_{\mathrm{RL}}$ \citep{chen2025pirl}. Varying $\sigma$ changes diversity and fidelity to the frozen policy; the Repeated-ODE comparison tests removal of integration noise.
\par\endgroup

\begin{table}[ht!]
\centering
\caption{Hyperparameters for Action Proposer in real-world and simulation environments, detailing the SDE noise level $\sigma$, action candidate counts $N_c$, and action horizon $H$. Both DA-Vanilla and DA-ABL share the same parameters. GPC-RANK utilizes its own generative sampling mechanism rather than SDE sampling, hence the SDE noise level $\sigma$ is not applicable (N/A).}
\label{tab:hyperparameters}
\footnotesize
\begin{tabular}{lllccc}
\toprule
Environment & Method & Task & $\sigma$ & $N_c$ & $H$\\
\midrule
\multirow{5}{*}{Real-World} & \multirow{4}{*}{\da} & Cup Sleeving & 0.1 & \multirow{4}{*}{8} & \multirow{4}{*}{50}\\
& & Charger Plugging & 0.05 & & \\
& & Cap Opening & 0.08 & & \\
& & Screw Insertion & 0.05 & & \\
\cmidrule{2-6}
& GPC-RANK & All Tasks & N/A & 8 & 50\\
\midrule
\multirow{4}{*}{Simulation} & \multirow{3}{*}{\da} & LIBERO Object \& Goal & 0.3 & \multirow{3}{*}{4} & 10\\
& & LIBERO Spatial \& 10 & 0.5 & &10 \\
& & SimplerEnv & 0.3 & & 16\\
\cmidrule{2-6}
& \multirow{2}{*}{GPC-RANK} & LIBERO & \multirow{2}{*}{N/A} & \multirow{2}{*}{4} & 10\\
&  & SimplerEnv &  &  & 16\\
\bottomrule
\end{tabular}
\end{table}

\subsection{Dream Evaluator}

\subsubsection{World Model} \label{app:world_model}

The world model is initialized with DreamDojo-2B \citep{gao2026dreamdojo} and post-trained on target robot data. To support low-latency online intervention, we follow the Self Forcing paradigm \citep{huang2025self} to further distill the bidirectional teacher world model into an autoregressive student model. This distillation process consists of two stages: (1) Warmup: under a teacher forcing mechanism, the student model's predictions are regressed onto the ODE trajectories generated by the teacher. (2) Distillation: utilizing a distribution matching method, the student model is trained autoregressively on its own generated context. This student model replaces the bidirectional attention mechanism with a causal attention mechanism and reduces the number of diffusion sampling steps, thereby achieving efficient online rollout generation while maintaining temporal consistency and action controllability. A single 13-frame video inference takes approximately 2 s on a single RTX 4090 GPU.

\subsubsection{Value Model} \label{app:value_model}

We regress the continuous progress delta within the execution horizon $H$:
\begin{equation}
z_t^{(k)} = \eta_{t+H} - \eta_t
\label{eq:value_target}
\end{equation}
where $\eta_t \in [0, 1]$ is the frame-level progress score provided by Robometer \cite{liang2026robometer}. This continuous label intuitively reflects the true physical contribution of the action $u_t^{(k)}$ towards task completion.
To ensure the network can unambiguously identify catastrophic consequences, we introduce a boundary penalty mechanism. If Robometer predicts an irreversible failure within the horizon (e.g., a dropping object or severe collision causing a precipitous drop in success probability $\mu_t$), we forcefully truncate and reset its target label $z_t^{(k)}$ to $-1$.
The final layer of the value model directly outputs the predicted progress delta $\hat{v}_t^{(k)}$ via a linear layer. Since the target label $z_t^{(k)}$ is strictly bounded within the $[-1, 1]$ range, the network implicitly learns to constrain its unnormalized outputs to a similar range during training.

\begingroup
\subsubsection{Failure Audit} \label{app:failure_audit}
We use synchronized video and proprioceptive logs to review 44 failed trials of DA-ABL. Two trials, one Charger Plugging and one Screw Insertion, have no timely intervention because the trigger misses the model-annotated pre-contact window or activates too late. They are excluded from the denominator for evaluating a scored failure-causing action.
For each of the remaining 42 trials, we attribute failure to the most recent evaluated action chunk whose executed interval directly leads to the first irreversible state transition. Six of these chunks score above the threshold, yielding a false-negative rate of 6/42 = 14.3\% and hazard-detection recall of 36/42 = 85.7\% (Wilson 95\% CI [72.2, 93.3]).

\begin{table}[ht!]

\centering\scriptsize
\caption{The six optimistic errors in the value-model audit. All were executed and subsequently caused failure.}
\label{tab:optimistic_errors}
\begin{tabularx}{\linewidth}{l c X}
\toprule
Task & Count & Observed failure mode\\
\midrule
Cup Sleeving & 1 & Contact with the sleeve rim\\
Charger Plugging & 2 & Socket-edge collision or lateral misalignment\\
Cap Opening & 1 & Gripper slippage\\
Screw Insertion & 2 & Subtle tip-to-hole misalignment\\
\bottomrule
\end{tabularx}
\end{table}

Most missed hazards involve fine contact geometry or occlusion insufficiently represented in the predicted video; two become apparent only near or beyond the prediction horizon. In the other 36 evaluated failures, the selected action is already below the diagnostic threshold, but is executed because it ranks highest. This exposes a limitation of relative ranking: a recognized hazardous action can still be the best available candidate. This recall measures recognition among evaluated failures.
\par\endgroup

\section{Real-World Evaluation Details} \label{app:real_world}

\subsection{Real-World Deployment Setup} \label{app:arm}

We deploy our base policy $\pi_{0.5}$ on a single NVIDIA RTX 4090 GPU. Table \ref{tab:task_info} details the task specifications, training steps (on 8 NVIDIA A100 GPUs), and demonstration data statistics (episodes and average lengths) for the real-world post-training of $\pi_{0.5}$.

\begin{table}[ht!]
\caption{Detailed task information for the real-world evaluation tasks. All videos in the teleoperation dataset are recorded at 30 FPS.}
\label{tab:task_info}
\centering
\footnotesize
\begin{tabularx}{\linewidth}{lXrrr}
\toprule
Task Name & Task Instruction & \shortstack{$\pi_{0.5}$ Training\\Steps} & \shortstack{Teleop\\Episodes} & \shortstack{Avg.\\Frames} \\
\midrule
Cup Sleeving & Pick up the paper cup and put it into the cup sleeve. & 40000 & 100 & 330.8 \\
Charger Plugging & Plug the charger into the socket. & 50000 & 100 & 336.5 \\
Cap Opening & Open the cap of the spray. & 50000 & 100 & 263.6 \\
Screw Insertion & Insert the screw into the hole in the box. & 50000 & 130 & 313.6 \\
\bottomrule
\end{tabularx}
\end{table}

\subsection{Multi-View Input Observations for Base Policy} \label{app:all_tasks_3view_sub}

\begin{figure*}[ht!]
    \centering
    \begin{subfigure}{0.8\textwidth}
        \centering
        \includegraphics[width=\textwidth]{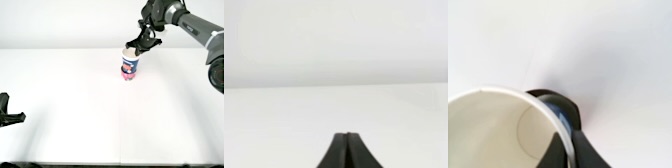}
        \caption{Cup Sleeving}
    \end{subfigure}
    
    \vspace{0.5em}
    
    \begin{subfigure}{0.8\textwidth}
        \centering
        \includegraphics[width=\textwidth]{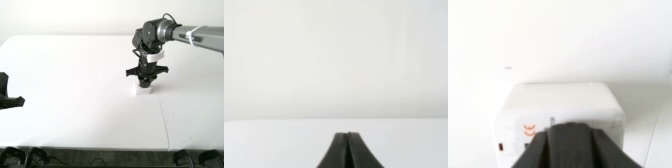}
        \caption{Charger Plugging}
    \end{subfigure}
    
    \vspace{0.5em}
    
    \begin{subfigure}{0.8\textwidth}
        \centering
        \includegraphics[width=\textwidth]{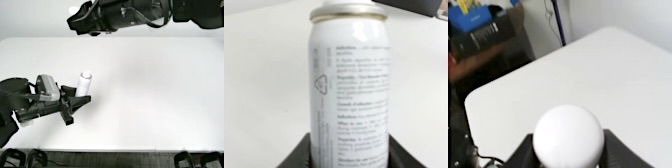}
        \caption{Cap Opening}
    \end{subfigure}
    
    \vspace{0.5em}
    
    \begin{subfigure}{0.8\textwidth}
        \centering
        \includegraphics[width=\textwidth]{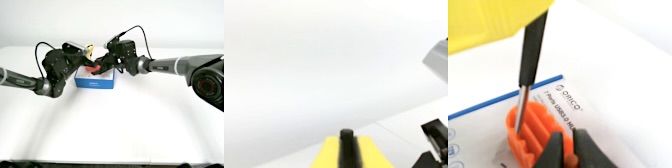}
        \caption{Screw Insertion}
    \end{subfigure}
    \caption{Multi-view input observations for base policy $\pi_{0.5}$ in real-world tasks (main view | left wrist view | right wrist view). Each row presents three synchronized camera views for a specific task, highlighting the need for accurate spatial alignment and contact establishment. These tasks are sensitive to minor action deviations compared to coarse-grained pick-and-place scenarios.}
    \label{fig:all_tasks_3view_sub}
\end{figure*}

Figure \ref{fig:all_tasks_3view_sub} shows three-view input observations for base policy $\pi_{0.5}$ in real-world tasks.

\subsection{Latency Analysis and Smooth Transition} \label{app:time_consumption}

\begingroup
Table~\ref{tab:time_consumption} compares latency at the trajectory
level under the deployed scheduling policies. \da invokes
the evaluator only at 1--3 detected critical phases, whereas GPC-RANK
replans every 25 frames. Over the observed
300--480-frame rollouts, \da requires an estimated
$2.69$--$7.33$\,s of cumulative inference, compared with
$8.33$--$13.88$\,s for GPC-RANK.
\par\endgroup

\begingroup
\begin{table}[ht!]

\centering\scriptsize
\caption{Measured per-call costs and estimated cumulative inference latency under the observed rollout schedules ($T=300$--$480$ frames).}
\label{tab:time_consumption}
\begin{tabularx}{\linewidth}{@{}l X c r@{}}
\toprule
Method & Deployment schedule & Calls / trajectory & 
Cumulative latency (s)\\
\midrule
Base $\pi_{0.5}$
    & Replan every 50 frames ($0.093$\,s/call)
    & 6--10
    & $0.56$--$0.93$\\

\textbf{\da (ours)}
    & Base replanning every 50 frames; intervention only at detected critical phases ($2.133$\,s/intervention)
    & 6--10 base + 1--3 interventions
    & $\mathbf{2.69}$--$\mathbf{7.33}$\\

GPC-RANK
    & Always-on replanning every 25 frames ($0.694$\,s/replan)
    & 12--20
    & $8.33$--$13.88$\\
\bottomrule
\end{tabularx}

\vspace{2pt}
\parbox{\linewidth}{\footnotesize
\emph{Per-call breakdown.}
A \da intervention comprises action proposal
($0.093$\,s), video generation ($2.000$\,s), and value scoring
($0.040$\,s). A GPC-RANK replan comprises candidate sampling
($0.093$\,s), world rollout ($0.600$\,s), and reward prediction
($0.001$\,s). The asynchronous \da trigger ($0.050$\,s/poll) runs
in a separate process and is excluded from the critical path.
}
\end{table}

\paragraph{Call Counts and Cumulative Inference.}
Observed evaluation rollouts contain approximately 300--480 frames, whereas the 260--340-frame range in the demonstration table describes training videos. At these rollout lengths, the base policy replans $\lceil T/50\rceil=6$--10 times and GPC-RANK replans $\lceil T/25\rceil=12$--20 times. \da triggers 1--3 interventions per trajectory. Using the measured mean per-call costs, a conservative additive accounting is
\begin{align}
L_{\mathrm{base}}(T)&=0.093\lceil T/50\rceil,\nonumber\\
L_{\mathrm{DA}}(T,m)&\approx0.093\lceil T/50\rceil+2.133m,\nonumber\\
L_{\mathrm{GPC}}(T)&=0.694\lceil T/25\rceil,
\label{eq:latency_count}
\end{align}
in seconds, with $m$ the intervention count. This yields 0.56--0.93\,s for the base policy, 2.69--7.33\,s for \da, and 8.33--13.88\,s for GPC-RANK across the stated ranges. The additive \da estimate can overcount a base-policy call when candidate sampling replaces that call. These are call-count estimates of cumulative inference, not measurements of end-to-end task duration.
The scheduling-dependent evaluator overhead is $mC_{\mathrm{eval}}$ rather than $\lceil T/r\rceil C_{\mathrm{eval}}$ for an always-on schedule with replan interval $r$.

\par\endgroup

\begingroup
\paragraph{Holding and Resuming Execution.}
During dreaming, high-level ROS command publication pauses while the low-level servos track the last position setpoint. After inference, the action sequence is reinitialized using the latest proprioceptive state; linear trajectory blending maintains $C^0$ command continuity. Resetting the high-level timer and adding a stabilization buffer prevents accumulated timer deficits from releasing commands in a burst. These implementation measures reduce command discontinuities.
\par\endgroup

\section{Simulation Benchmark Details} \label{app:simulation}

\begin{figure*}[ht!]
  \centering
  \includegraphics[width=\linewidth]{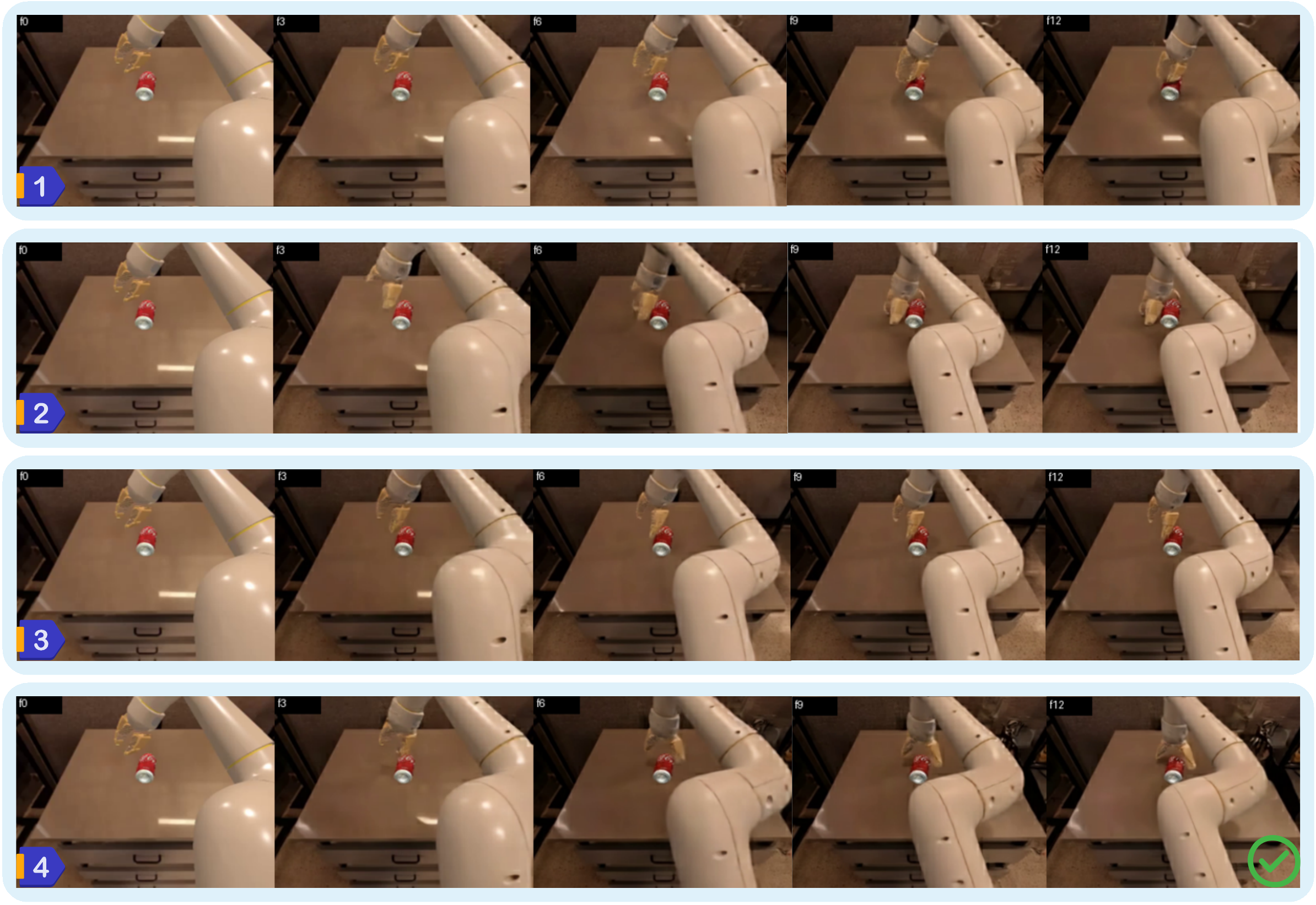}
  \caption{On the google\_robot\_pick\_coke\_can task of the SimplerEnv benchmark, future videos are generated by DreamDojo based on 4 action chunk candidates produced by GR00T-N1.6 via SDE sampling (13 frames in total; frames 0, 3, 6, 9, and 12 are shown here). Among these, (4) has the highest corresponding value score.} 
    \label{fig:simplerenv_4}
\end{figure*}

\begingroup
Simulation follows the real-world annotation and training procedure: Gemini 3.6 Flash annotates critical-phase onsets offline from training videos and task instructions. These annotations supervise a lightweight Dream Trigger that schedules interventions using only causal visual observations and benchmark-provided proprioception at test time. Gemini 3.6 Flash is used solely for offline annotation. Trigger and SDE sampling parameters are selected on a separate validation split and fixed before testing. GPC-RANK is always-on.

Tables~\ref{tab:libero} and~\ref{tab:simplerenv} report success rates and Wilson 95\% intervals. The accounting is 200 episodes for each of the four displayed LIBERO settings (800 total), and 200 per SimplerEnv task (1400 Bridge and 1000 Fractal episodes). The intervals and tests are computed from these aggregate counts. Figure~\ref{fig:simplerenv_4} illustrates four predicted futures.
\par\endgroup

\begin{table}[ht!]
\centering\scriptsize\setlength{\tabcolsep}{4pt}
\caption{LIBERO success (\%) and Wilson 95\% intervals. Each displayed setting contains 200 episodes; the average pools 800 episodes.}
\label{tab:libero}
\begin{tabular}{lccccc}
\toprule
Method & Object & Goal & 10 & Spatial & Average\\
\midrule
Maximum steps & 280 & 300 & 520 & 220 & 330\\
\midrule
$\pi_{0.5}$ & \shortstack{98.0\\\scriptsize[95.0, 99.2]} & \shortstack{98.0\\\scriptsize[95.0, 99.2]} & \shortstack{91.0\\\scriptsize[86.2, 94.2]} & \shortstack{99.0\\\scriptsize[96.4, 99.7]} & \shortstack{96.5\\\scriptsize[95.0, 97.6]}\\
GPC-RANK & \shortstack{99.0\\\scriptsize[96.4, 99.7]} & \shortstack{97.5\\\scriptsize[94.3, 98.9]} & \shortstack{92.0\\\scriptsize[87.4, 95.0]} & \shortstack{98.0\\\scriptsize[95.0, 99.2]} & \shortstack{96.6\\\scriptsize[95.1, 97.7]}\\
DA-Vanilla & \shortstack{98.5\\\scriptsize[95.7, 99.5]} & \shortstack{97.0\\\scriptsize[93.6, 98.6]} & \shortstack{93.0\\\scriptsize[88.6, 95.8]} & \shortstack{98.5\\\scriptsize[95.7, 99.5]} & \shortstack{96.8\\\scriptsize[95.3, 97.8]}\\
DA-ABL & \shortstack{98.5\\\scriptsize[95.7, 99.5]} & \shortstack{99.0\\\scriptsize[96.4, 99.7]} & \shortstack{94.5\\\scriptsize[90.4, 96.9]} & \shortstack{99.0\\\scriptsize[96.4, 99.7]} & \shortstack{97.8\\\scriptsize[96.5, 98.6]}\\
\bottomrule\end{tabular}\end{table}

\begin{table}[ht!]
\centering\scriptsize\setlength{\tabcolsep}{4pt}
\caption{SimplerEnv success (\%) and Wilson 95\% intervals, with 200 episodes per task. Averages pool 1400 Bridge or 1000 Fractal episodes. Task labels are shortened for readability.}
\label{tab:simplerenv}
\begin{tabular}{lcccc}
\toprule
Task & GR00T-N1.6 & GPC-RANK & DA-Vanilla & DA-ABL\\
\midrule
\multicolumn{5}{l}{Bridge (WidowX robot)}\\
\midrule
Spoon on towel & \shortstack{62.5\\\scriptsize[55.6, 68.9]} & \shortstack{64.5\\\scriptsize[57.7, 70.8]} & \shortstack{65.0\\\scriptsize[58.2, 71.3]} & \shortstack{67.0\\\scriptsize[60.2, 73.1]}\\
Carrot on plate & \shortstack{63.0\\\scriptsize[56.1, 69.4]} & \shortstack{64.0\\\scriptsize[57.1, 70.3]} & \shortstack{64.0\\\scriptsize[57.1, 70.3]} & \shortstack{65.5\\\scriptsize[58.7, 71.7]}\\
Eggplant in basket & \shortstack{89.5\\\scriptsize[84.5, 93.0]} & \shortstack{91.0\\\scriptsize[86.2, 94.2]} & \shortstack{92.5\\\scriptsize[88.0, 95.4]} & \shortstack{95.0\\\scriptsize[91.0, 97.3]}\\
Stack cube & \shortstack{4.0\\\scriptsize[2.0, 7.7]} & \shortstack{6.0\\\scriptsize[3.5, 10.2]} & \shortstack{5.5\\\scriptsize[3.1, 9.6]} & \shortstack{7.0\\\scriptsize[4.2, 11.4]}\\
Eggplant in sink & \shortstack{38.0\\\scriptsize[31.6, 44.9]} & \shortstack{38.0\\\scriptsize[31.6, 44.9]} & \shortstack{40.5\\\scriptsize[33.9, 47.4]} & \shortstack{41.0\\\scriptsize[34.4, 47.9]}\\
Close drawer & \shortstack{68.0\\\scriptsize[61.2, 74.1]} & \shortstack{70.5\\\scriptsize[63.8, 76.4]} & \shortstack{70.0\\\scriptsize[63.3, 75.9]} & \shortstack{73.5\\\scriptsize[67.0, 79.1]}\\
Open drawer & \shortstack{94.0\\\scriptsize[89.8, 96.5]} & \shortstack{94.0\\\scriptsize[89.8, 96.5]} & \shortstack{95.5\\\scriptsize[91.7, 97.6]} & \shortstack{96.0\\\scriptsize[92.3, 98.0]}\\
\midrule Average & \shortstack{59.9\\\scriptsize[57.3, 62.4]} & \shortstack{61.1\\\scriptsize[58.6, 63.7]} & \shortstack{61.9\\\scriptsize[59.3, 64.4]} & \shortstack{63.6\\\scriptsize[61.0, 66.1]}\\
\midrule\multicolumn{5}{l}{Fractal (Google robot)}\\\midrule
Pick coke can & \shortstack{97.0\\\scriptsize[93.6, 98.6]} & \shortstack{97.5\\\scriptsize[94.3, 98.9]} & \shortstack{97.0\\\scriptsize[93.6, 98.6]} & \shortstack{98.5\\\scriptsize[95.7, 99.5]}\\
Pick object & \shortstack{85.0\\\scriptsize[79.4, 89.3]} & \shortstack{87.0\\\scriptsize[81.6, 91.0]} & \shortstack{88.0\\\scriptsize[82.8, 91.8]} & \shortstack{89.5\\\scriptsize[84.5, 93.0]}\\
Move near & \shortstack{73.5\\\scriptsize[67.0, 79.1]} & \shortstack{74.0\\\scriptsize[67.5, 79.6]} & \shortstack{76.0\\\scriptsize[69.6, 81.4]} & \shortstack{78.5\\\scriptsize[72.3, 83.6]}\\
Open drawer & \shortstack{44.0\\\scriptsize[37.3, 50.9]} & \shortstack{45.0\\\scriptsize[38.3, 51.9]} & \shortstack{48.5\\\scriptsize[41.7, 55.4]} & \shortstack{51.0\\\scriptsize[44.1, 57.8]}\\
Close drawer & \shortstack{82.5\\\scriptsize[76.6, 87.1]} & \shortstack{87.5\\\scriptsize[82.2, 91.4]} & \shortstack{85.0\\\scriptsize[79.4, 89.3]} & \shortstack{86.0\\\scriptsize[80.5, 90.1]}\\
\midrule Average & \shortstack{76.4\\\scriptsize[73.7, 78.9]} & \shortstack{78.2\\\scriptsize[75.5, 80.6]} & \shortstack{78.9\\\scriptsize[76.3, 81.3]} & \shortstack{80.7\\\scriptsize[78.1, 83.0]}\\
\bottomrule\end{tabular}\end{table}

\begingroup
RoboTwin 2.0 uses the Aloha-AgileX bimanual robot. We evaluate eight tasks with 100 episodes per task in the demo\_clean setting. All methods re-plan every 25 steps. Table~\ref{tab:robotwin} reports success rates and Wilson 95\% intervals; the CMH test against the receding-horizon $\pi_{0.5}$ gives $p=0.634$ for fine-tuning on demonstrations alone and $p=0.028$ with rollouts.
\par\endgroup

\begin{table}[ht!]
\centering\scriptsize\setlength{\tabcolsep}{4pt}
\caption{RoboTwin 2.0 success (\%) and Wilson 95\% intervals, 100 episodes per task.}
\label{tab:robotwin}
\begin{tabular}{lcc}
\toprule
Task & $\pi_{0.5}$ & DA-ABL \\
\midrule
Hanging mug & \shortstack{30.0\\{\scriptsize[21.9, 39.6]}} & \shortstack{19.0\\{\scriptsize[12.5, 27.8]}}\\
Handover block & \shortstack{82.0\\{\scriptsize[73.3, 88.3]}} & \shortstack{94.0\\{\scriptsize[87.5, 97.2]}}\\
Stack bowls (3) & \shortstack{77.0\\{\scriptsize[67.8, 84.2]}} & \shortstack{74.0\\{\scriptsize[64.6, 81.6]}}\\
Stack blocks (3) & \shortstack{96.0\\{\scriptsize[90.2, 98.4]}} & \shortstack{98.0\\{\scriptsize[93.0, 99.4]}}\\
Place phone stand & \shortstack{75.0\\{\scriptsize[65.7, 82.5]}} & \shortstack{78.0\\{\scriptsize[68.9, 85.0]}}\\
Stamp seal & \shortstack{63.0\\{\scriptsize[53.2, 71.8]}} & \shortstack{79.0\\{\scriptsize[70.0, 85.8]}}\\
Place can in basket & \shortstack{74.0\\{\scriptsize[64.6, 81.6]}} & \shortstack{73.0\\{\scriptsize[63.6, 80.7]}}\\
Move stapler to pad & \shortstack{33.0\\{\scriptsize[24.6, 42.7]}} & \shortstack{51.0\\{\scriptsize[41.3, 60.6]}}\\
\midrule
Average & \shortstack{66.3\\{\scriptsize[62.9, 69.4]}} & \shortstack{70.8\\{\scriptsize[67.5, 73.8]}}\\
\bottomrule
\end{tabular}
\end{table}

\begingroup
\section{Statistical Reporting} \label{app:statistics}

Table~\ref{tab:real_summary} summarizes aggregate real-world results across four tasks, with 40 trials per task per method. Table~\ref{tab:real_intervals} provides per-task success rates and confidence intervals, including ablations.

\begingroup
\begin{table}[ht!]
\centering\scriptsize
\caption{Real-world success over 160 trials per method. Confidence intervals are Wilson 95\%; CMH tests compare each method with DA-ABL while stratifying by task.}
\label{tab:real_summary}
\begin{tabular}{lccc}
\toprule
Method & Successes & Success (\%) [95\% CI] & $p$ vs.\ DA-ABL\\
\midrule
Base $\pi_{0.5}$ & 78/160 & 48.8 [41.1, 56.4] & $7.4\times10^{-6}$\\
GPC-RANK & 87/160 & 54.4 [46.6, 61.9] & $5.3\times10^{-4}$\\
DA-Vanilla & 107/160 & 66.9 [59.3, 73.7] & 0.251\\
DA-ABL & 116/160 & 72.5 [65.1, 78.8] & ---\\
\bottomrule
\end{tabular}
\end{table}
\par\endgroup

\paragraph{Counts and Confidence Intervals.}
Real-world rates use 40 trials per task per method. Figure~\ref{fig:result}'s base-policy labels are rendered at one decimal place consistently with the underlying counts: 62.5\% for Cup Sleeving, 47.5\% for Charger Plugging, and 48.8\% overall. Table~\ref{tab:real_intervals} includes all methods and ablations.
For $k$ successes among $n$ trials, $\hat p=k/n$ and $z_{.975}=1.959964$, the Wilson interval is
\begin{equation}
\frac{\hat p+z_{.975}^2/(2n)\ \pm\ z_{.975}\sqrt{\hat p(1-\hat p)/n+z_{.975}^2/(4n^2)}}{1+z_{.975}^2/n}.
\end{equation}
Aggregate intervals describe the pooled episode success rate across the fixed evaluated tasks.

\begin{table}[ht!]

\centering\scriptsize\setlength{\tabcolsep}{4pt}
\caption{Real-world success (\%) with Wilson 95\% intervals, 40 trials per cell.}
\label{tab:real_intervals}
\begin{tabular}{lcccc}
\toprule
Method & Cup & Charger & Cap & Screw\\
\midrule
Base $\pi_{0.5}$ & \shortstack{62.5\\\scriptsize[47.0, 75.8]} & \shortstack{47.5\\\scriptsize[32.9, 62.5]} & \shortstack{55.0\\\scriptsize[39.8, 69.3]} & \shortstack{30.0\\\scriptsize[18.1, 45.4]}\\
GPC-RANK & \shortstack{60.0\\\scriptsize[44.6, 73.7]} & \shortstack{55.0\\\scriptsize[39.8, 69.3]} & \shortstack{67.5\\\scriptsize[52.0, 79.9]} & \shortstack{35.0\\\scriptsize[22.1, 50.5]}\\
DA-Vanilla & \shortstack{82.5\\\scriptsize[68.1, 91.3]} & \shortstack{60.0\\\scriptsize[44.6, 73.7]} & \shortstack{80.0\\\scriptsize[65.2, 89.5]} & \shortstack{45.0\\\scriptsize[30.7, 60.2]}\\
Repeated ODE & \shortstack{67.5\\\scriptsize[52.0, 79.9]} & \shortstack{55.0\\\scriptsize[39.8, 69.3]} & \shortstack{70.0\\\scriptsize[54.6, 81.9]} & \shortstack{37.5\\\scriptsize[24.2, 53.0]}\\
Random Select & \shortstack{65.0\\\scriptsize[49.5, 77.9]} & \shortstack{35.0\\\scriptsize[22.1, 50.5]} & \shortstack{55.0\\\scriptsize[39.8, 69.3]} & \shortstack{25.0\\\scriptsize[14.2, 40.2]}\\
DA-ABL & \shortstack{90.0\\\scriptsize[76.9, 96.0]} & \shortstack{67.5\\\scriptsize[52.0, 79.9]} & \shortstack{80.0\\\scriptsize[65.2, 89.5]} & \shortstack{52.5\\\scriptsize[37.5, 67.1]}\\
\bottomrule\end{tabular}\end{table}

\paragraph{Comparative Tests.}
We use a two-sided Cochran--Mantel--Haenszel (CMH) test without continuity correction, stratifying by the reported task or evaluation setting. Each stratum is a $2\times2$ table of method by success/failure. This treats episodes as the statistical units and does not assume matched trial pairs.
The main-text difference intervals are unpaired normal-approximation intervals computed as
\begin{equation}
(\hat p_1-\hat p_0)\ \pm\ z_{.975}\sqrt{\frac{\hat p_1(1-\hat p_1)}{n_1}+\frac{\hat p_0(1-\hat p_0)}{n_0}}.
\end{equation}
These estimate the aggregate risk difference. Real-world DA-ABL versus base and GPC-RANK gives $p=7.3848\times10^{-6}$ and $5.2979\times10^{-4}$, respectively. DA-ABL versus DA-Vanilla gives $p=0.2512$; versus Repeated ODE, $p=0.003616$; and versus Random Select, $p=1.9048\times10^{-7}$. The last two values use the same CMH procedure as the main comparisons. For simulation, DA-ABL versus base gives $p=0.1306$ on LIBERO, $0.01225$ on Bridge, and $0.01015$ on Fractal; on RoboTwin, the rollout-fine-tuned $\pi_{0.5}$ versus the receding-horizon $\pi_{0.5}$ gives $p=0.02761$. Statistical significance here concerns the reported episode outcomes.
\par\endgroup

\section{\texorpdfstring{Limitations and Future Work}{Limitations and Future Work}} \label{app:limitation}

First, although \da strictly limits dense computations to sparse critical phases, generating high-fidelity future videos and scoring multiple candidate actions still introduces latency. For high-precision but slower-paced manipulation tasks, this overhead can be acceptable. However, in highly dynamic environments requiring split-second reactive control, this latency may hinder the robot's smooth execution. In addition to employing various video generation acceleration techniques, future research could also perform forward consequence simulation within a compact latent space. It must be emphasized, however, that explicitly generating pixel-level videos holds practical value: its intuitive interpretability not only enables human researchers to quickly diagnose and optimize the model, but these synthesized failure scenarios and corrective trajectories can also be validated for use as negative samples or preference data for the continuous fine-tuning of subsequent VLA or world models (e.g., via DPO or RLHF).

Secondly, although this paper primarily uses \da as an enhancement module for VLA models, its core test-time dreaming mechanism is modular with respect to the underlying base policy. Future work could integrate this failure-aware framework as a test-time intervention layer into broader embodied control architectures, such as the emerging World Action Models or more traditional control policies. For specific implementations, adaptive modifications to the action sampling methods of different architectures are required, together with task-specific trigger training and evaluator adaptation: for instance, SDE sampling can be adopted for action trajectory search in flow matching-based policies, while diverse candidate actions can be generated by injecting Gaussian perturbations in diffusion-based models.

\begingroup
\section{Broader Impacts} \label{app:broad}
Improved empirical reliability in contact-sensitive manipulation could reduce some failed executions and wasted robot interaction. The present evidence does not establish safety in homes, factories, or out-of-distribution scenes, and the system may still select hazardous actions or pause in unfavorable contact states. Appropriate human supervision and independent hardware safeguards remain necessary. More reliable automation can also affect workers performing related manual tasks; deployment should consider these social consequences alongside task performance.
\par\endgroup

\end{document}